\documentclass[sigconf,natbib=true,screen=true]{acmart}
\usepackage{graphicx}
\usepackage{soul}
\usepackage{amsmath}

\usepackage{amssymb}
\usepackage{amsthm}
\usepackage{bm}
\usepackage{booktabs}
\usepackage{multirow}
\usepackage[inline]{enumitem}

\usepackage{tabularx}
\usepackage{subfigure}
\usepackage{color, xcolor}

\usepackage{definitions}

\usepackage{balance}

\AtBeginDocument{%
  }

\copyrightyear{2023}
\acmYear{2023}
\setcopyright{rightsretained}
\acmConference[CIKM '23]{Proceedings of the 32nd ACM International Conference on Information and Knowledge Management}{October 21--25, 2023}{Birmingham, United Kingdom}
\acmBooktitle{Proceedings of the 32nd ACM International Conference on Information and Knowledge Management (CIKM '23), October 21--25, 2023, Birmingham, United Kingdom}
\acmDOI{10.1145/3583780.3614934}
\acmISBN{979-8-4007-0124-5/23/10}

\makeatletter
\g@addto@macro\normalsize{%
  \abovedisplayskip 0pt plus1pt 
  \belowdisplayskip 0pt plus1pt
  \abovedisplayshortskip  0pt plus1pt%
  \belowdisplayshortskip  0pt plus1pt
}
\makeatother

\begin{document}

\title{Inducing Causal Structure for Abstractive Text Summarization}

\author{Lu Chen}
\orcid{0000-0001-6747-8015}
\affiliation{
	\institution{CAS Key Lab of Network Data Science and Technology, ICT, CAS}
	\institution{University of Chinese Academy of Sciences}
	\city{Beijing}
	\country{China}
}
\email{chenlu19z@ict.ac.cn}

\author{Ruqing Zhang}
\authornote{Research conducted when the author was at the University of Amsterdam.}
\orcid{0000-0003-4294-2541}
\affiliation{
	\institution{CAS Key Lab of Network Data Science and Technology, ICT, CAS}
	\institution{University of Chinese Academy of Sciences}
	\city{Beijing}
	\country{China}
}
\email{zhangruqing@ict.ac.cn}

\author{Wei Huang}
\orcid{0009-0001-1556-8198}
\affiliation{
	\institution{CAS Key Lab of Network Data Science and Technology, ICT, CAS}
	\institution{University of Chinese Academy of Sciences}
 \city{Beijing}
 \country{China}
}
\email{huangwei21b@ict.ac.cn}
 
\author{Wei Chen}
\orcid{0000-0002-7438-5180}
\authornote{Wei Chen is the corresponding author.}
\affiliation{
	\institution{CAS Key Lab of Network Data Science and Technology, ICT, CAS}
	\institution{University of Chinese Academy of Sciences}
 \city{Beijing}
 \country{China}
}
\email{chenwei2022@ict.ac.cn}

\author{Jiafeng Guo}
\orcid{0000-0002-9509-8674}
\affiliation{
	\institution{CAS Key Lab of Network Data Science and Technology, ICT, CAS}
	\institution{University of Chinese Academy of Sciences}
	\city{Beijing}
	\country{China}
}
\email{guojiafeng@ict.ac.cn}

\author{Xueqi Cheng}
\orcid{0000-0002-5201-8195}
\affiliation{
	\institution{CAS Key Lab of Network Data Science and Technology, ICT, CAS}
	\institution{University of Chinese Academy of Sciences}
	\city{Beijing}
	\country{China}
}
\email{cxq@ict.ac.cn}

\renewcommand{\shortauthors}{Lu Chen et al.}

\begin{abstract}
    The mainstream of data-driven abstractive summarization models tends to explore the correlations rather than the causal relationships. 
    Among such correlations, there can be spurious ones which suffer from the language prior learned from the training corpus and therefore undermine the overall effectiveness of the learned model. 
    To tackle this issue, we introduce a Structural Causal Model (SCM) to induce the underlying causal structure of the summarization data.  
    We assume several latent causal factors and non-causal factors, representing the content and style of the document and summary. 
    Theoretically, we prove that the latent factors in our SCM can be identified by fitting the observed training data under certain conditions.
    On the basis of this, we propose a Causality Inspired Sequence-to-Sequence model (CI-Seq2Seq) to learn the causal representations that can mimic the causal factors, guiding us to pursue causal information for summary generation. 
    The key idea is to reformulate the Variational Auto-encoder (VAE) to fit the joint distribution of the document and summary variables from the training corpus. 
    Experimental results on two widely used text summarization datasets demonstrate the advantages of our approach. 
\end{abstract}




\maketitle

\section{Introduction}
Text summarization is an important task in natural language processing (NLP), which targets to produce a fluent and condensed summary for a document, while preserving the key information \cite{maybury1999advances,nenkova2012survey}.
Abstractive summarization is a mainstream approach to generate compact summaries from scratch \cite{banko2000headline,zajic2004bbn}.
Advances in deep learning have fueled research in applying neural sequence-to-sequence (Seq2Seq) networks to automatically extract effective features and generate summaries in an end-to-end manner \cite{rush2015neural,nallapati2017summarunner,paulus2018deep}.

Despite the promising performance, current data-driven summarization models possess an inherent issue.
These efforts often exploit all types of correlations to fit data well, overlooking the underlying data generating process (DGP) that reveals how observed data is generated  \cite{ahuja2022towards}.
Such correlations are probably spurious due to the biased statistical dependencies caused by confounder inherited from the training corpus. 
For instance, if the term ``lion'' frequently co-occurs with ``Africa'' in training data, a model might erroneously generate a summary containing ``Africa'' even for the document describing the core information ``lion pregnancy'' with the side information ``Africa''.
The occurrence of such stereotyping, arising from spurious correlations, impacts the effectiveness of text summarization techniques and hinders practical applications.

Recently, structural causal model (SCM) has attracted great interest from the research community to identify the underlying DGP of the observed data \cite{pearl2009causal,pearl2009causality,altman2015points,moraffah2020causal,scholkopf2022causality}. 
Learned causal models aid stable prediction and generalization by capturing causal relationships.
In this work, we aim to devise a SCM for describing the DGP in text summarization, with the goal of inducing causal structure of the data, especially the causal relationships between documents and summaries.  
We would like the latent space to be separated into content and style space. 
For the content space, we assume two kinds of latent factors, i.e., Core-Content (CC) factors and  Side-Content (SC) factors, referring to the core content (main points) and side content (non-essential information) in the document, respectively. 
For the style space, we also assume two kinds of latent factors, i.e., Document-Style (DS) factors and Summary-Style (SS) factors, referring to the lengthy writing style of the document and the concise writing style of the summary, respectively.   
Among such latent factors, there can be confounder representing the statistical dependencies inherited from the training corpus.

\begin{figure}[t]
  \centering
   \includegraphics[scale=0.44]{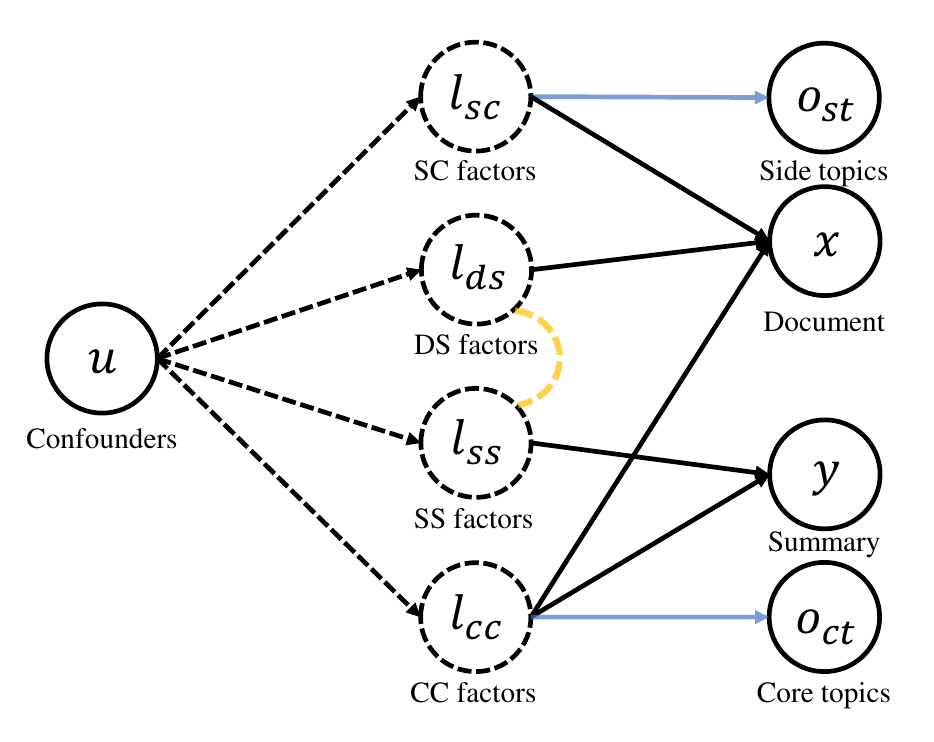}
   \caption{The proposed SCM for text summarization.
   Solid and dashed circles denote observed and latent variables.
   The solid arrows pointing to $x$ and $y$ represent the invariant causal generative mechanisms $p(x|l_{cc},l_{sc},l_{ds})$ and $p(y|l_{cc},l_{ss})$, while the dashed arrows pointing from $u$ represent the varied latent distributions given confounder.
   The blue arrows pointing to $o_{st}$ and $o_{ct}$ represent the content guidance for $l_{sc}$ and $l_{cc}$, while the yellow dashed line between $l_{ds}$ and $l_{ss}$ represents their relation as the style guidance. 
   Details see Section \ref{causal view}.}
   \label{fig:scm}
\end{figure}

Specifically, as shown in Figure \ref{fig:scm}, we assume that CC and SS factors are summary-causal factors whose relationship with the summary remains invariant across the corpus, while other factors are non-causal for the summary and only causally influence the document. 
Each document is generated from the summary-causal factor CC and the non-causal factors SC and DS. 
Besides, we incorporate core topics and side topics in the documents to guide the learning of CC and SC factors. 
Theoretically, we prove that certain conditions ensure the identifiability of causal factors, enabling the generation of summaries containing only causal information.

Based on the SCM, we propose a Causality Inspired Sequence-to-Sequence model (CI-Seq2Seq) for abstractive text summarization, which enforces the learned representations to mimic the latent factors. 
The key idea is to learn the \emph{causal generative mechanisms} for the document and summary, by adapting Variational Auto-encoder (VAE) \cite{kingma2013auto} to supervised training.
Specifically, we first partition each dataset into subsets through Latent Dirichlet Allocation (LDA) \cite{blei2003latent} and define confounder information as topical features of subsets. 
Then, we utilize LDA and Compression Rate (CR) \cite{hahn2000challenges} as the guidance to learn the content and style factors, respectively.
During testing, we first infer CC and DS factors based on the learned document-causal generative mechanisms, and then use the summary-causal generative mechanisms for controlled summary generation based on the given CR between DS and SS factors.

To the best of our knowledge, it is the first work to combine causality and text summarization with a rigorous theoretical guarantee.  
Experimental results on two widely-used datasets, i.e., CNN/Daily Mail \cite{hermann2015teaching} and XSUM \cite{narayan2018don}, demonstrated that CI-Seq2Seq  can achieve significant improvements over prevailing baselines in terms of prediction performance, generalizability and interpretability. 
The code is available at https://github.com/ict-bigdatalab/CI-Seq2Seq.

\vspace{-4mm}
\section{Related Work}

\textbf{Text Summarization}. Text summarization can be categorized into extractive and abstractive methods.
Extractive methods directly extract and rearrange sentences from the document to generate the summary  \cite{nallapati2017summarunner,zhou2018neural,narayan2018ranking,zhong2019searching}. 
Abstractive methods aim to generate a summary by comprehending the document \cite{rush2015neural,liu2015toward,paulus2018deep,gehrmann2018bottom,fan2018controllable}. 
Recently, researchers have explored to utilize pre-trained language models (PLMs) to enhance the performance of both extractive \cite{wang2019self,zhong2020extractive,xu2020unsupervised,ruan2022histruct+} and abstractive methods \cite{liu2018generative,zou2020pre,lewis-etal-2020-bart,chen2021simple}. 
However, most studies capture only correlations rather than causal relationships.

\noindent \textbf{Causality for NLP}. Causality targets to explore the causal relationships in the data  \cite{pearl2009causality,yao2021survey}, which has been widely studied in various tasks, e.g., information retrieval \cite{joachims2017unbiased}, recommendation  \cite{zheng2021disentangling,zhang2021causal,wei2021model} and computer vision \cite{niu2021counterfactual,sun2021recovering,zhang2021deep}.
The mainstream methods include potential outcome model \cite{rubin1974estimating} and structural causal model (SCM) \cite{pearl1995causal,pearl2000models}.
Specifically, SCM has two primary applications: causal inference and causal representation learning. The former explores variable impact via causal intervention \cite{glymour2016causal} and counterfactual reasoning \cite{bottou2013counterfactual}.
The latter identifies causal factors by studying data generating process, which enhances robustness and generalizability, even when facing distributional shifts \cite{lu2021invariant,wang2022causal,mitrovicrepresentation,liu2021learning,zhang2021deep}.

As for NLP, the causality-aware methods are mainly studied in text classification \cite{qian2021counterfactual,veitch2021counterfactual}, table-to-text generation \cite{chen2021confounded} and language model pre-training \cite{cao2022can}, for debiasing \cite{qian2021counterfactual}, controlling \cite{hu2021causal} or style transferring  \cite{nangi2021counterfactuals}.
For example, some works  \cite{hu2021causal,chen2021confounded,cao2022can} applied causal intervention to eliminate the spurious correlations introduced by backdoor path. 
Some works \cite{chen2020exploring,xie2021factual,qian2021counterfactual,hu2021causal,nangi2021counterfactuals} utilized counterfactual reasoning to measure the causal effect by excluding the direct effect from its total effect, or control textual attributes by assigning counterfactual values. 
Yet there have been few works that apply causal perspective to text summarization \cite{xie2021factual}, particularly in terms of causal representation learning.

\noindent \textbf{Disentangled Representation Learning}. Disentangled representation learning aims to map different aspects of data into distinct low-dimensional latent spaces.
It has attracted considerable attention in machine learning \cite{hsu2017unsupervised} and NLP \cite{zou2022divide,dougrez2021learning,zeng2019you,nangi2021counterfactuals}.  
Besides disentangling latent factors, we focus on characterizing the causal and non-causal factors for text summarization.

\vspace{-2mm}
\section{A Causal View on Summarization}
\label{causal view} 

Following the definition that \textit{two variables have a causal relationship, denoted as ``cause $\to$ effect'', if intervening the cause may alter the effect, but not vice versa} \cite{pearl2009causality,peters2017elements}, we first define the causal relationships in text summarization, and then formulate it using structural causal model (SCM) \cite{pearl2000models}, followed by the identifiability analysis to ensure that the latent factors in our SCM can be correctly separated and learned under certain conditions.

\vspace{-2mm}
\subsection{Causal Relationships}

We introduce step-by-step about how we characterize the causal relationships in text summarization.

\noindent\textbf{(1) Assuming latent factors with causal relationships}. 
It is likely that there exist correlations between a document $x$ and its summary $y$. According to the Reichenbach’s common cause principle \cite{peters2017elements}, correlations mean there exist common causes that causally influence $x$ and $y$. 
We assume latent factors $z$ as common causes, carrying mixed information of $x$ and $y$.
That is, $z \to x$ and $z \to y$.

\noindent\textbf{(2) Clarifying the causes for document and summary}.
To separate the mixed information in $z$, we decompose $z$ in terms of content and style, i.e., Core-Content (CC) factor $l_{cc}$ and Side-Content (SC) factor $l_{sc}$ in content space, as well as Document-Style (DS) factor $l_{ds}$ and Summary-Style (SS) factor $l_{ss}$ in style space. 
For content, summary $y$ should preserve the core content while omit the side content of document $x$, i.e., $l_{cc} \to y$, $l_{cc} \to x$ and $l_{sc} \to x$. 
For style, considering different style of $x$ and $y$, we assume $l_{ds}$ is the style factor for $x$ and $l_{ss}$ for $y$, i.e., $l_{ds} \to x$ and $l_{ss} \to y$.

\noindent\textbf{(3) Capturing correlations among latent factors}. Latent factors may mix through spurious correlation from biased statistical dependencies of the training corpus \cite{qian2021counterfactual,nan2021uncovering}. We use $u$ to denote confounder resulting in the spurious correlation, and orient four edges from $u$ to latent factors, i.e., $u \to l_{cc},u \to l_{sc},u \to l_{ds},u \to l_{ss}$.

\noindent\textbf{(4) Adding guidance to separate latent factors}.
Practically, we use weakly-supervised signals to guide latent factor learning. 
For content factors, we introduce core topics $o_{ct}$ and side topics $o_{st}$ for $l_{cc}$ and $l_{sc}$ respectively.
That is, $l_{cc} \to o_{ct}$ and $l_{sc} \to o_{st}$.
For style factors, we define the function relation between $l_{ds}$ and $l_{ss}$ to bridge them.
Thus, we link $l_{ds}-l_{ss}$ by an undirected edge.

\vspace{-2mm}
\subsection{Structural Causal Model}
\label{section:SCM}
Based on the above analysis, we devise the SCM for text summarization (Figure \ref{fig:scm}). It describes the data generating process (DGP) – latent factors generate the observations (document and summary) given the confounder.
The nodes denote variables, and the edges denote relationships (directed: causal, undirected: non-causal).

We refer to $p(x|l_{cc},l_{sc},l_{ds})$ and $p(y|l_{cc},l_{ss})$ as the causal generative mechanisms for the document and summary, respectively.
They are assumed to be invariant to the prior $p(l_{cc},l_{sc},l_{ds},l_{ss})$ according to the Independent Causal Mechanisms (ICM) Principle \cite{ScholkopfJPSZM12,peters2017elements}, denoted by solid arrows in Figure \ref{fig:scm}, while the latent distributions given the confounder may vary across domains, denoted by dashed arrows.
Besides, the topic distributions $p(o_{ct}|l_{cc})$ and $p(o_{st}|l_{sc})$ denote the content guidance, and the function relation between $l_{ds}$ and $l_{ss}$ denotes the style guidance.
We formally present a comprehensive functional form for the DGP as outlined below.

We define $\Theta \triangleq \{f, {\Phib}\}$ as the parameters to generate observed variables, where $f$ is the invertible function mapping latent factors to observed variables, and ${\Phib}$ denotes the parameters to generate latent factor given confounder $u$.
The parent set is denoted as $Pa(\cdot)$.

Taking $x$ as an example, $Pa(x) = \{l_{sc}, l_{ds}, l_{cc}\}$. 
The joint probability density of $x$ and $Pa(x)$ can be written as:
\begin{align}
    \label{eq:gen}
    p_{\Theta_x}(x, Pa(x)|u) = &p_{\Theta_x}(x, l_{sc}, l_{cc}, l_{ds} |u)  \nonumber \\
    =&p_{f_x}(x|l_{sc}, l_{cc}, l_{ds})\cdot p_{{\Phib}_x}(l_{sc}, l_{cc}, l_{ds}|u).
\end{align}

For $p_{{\Phib}_x}(l_{sc}, l_{cc}, l_{ds}|u)$, we assume it follows exponential family:
\begin{align}
   \resizebox{1\hsize}{!}{$p_{{\Phib}_x}(l_{sc}, l_{cc}, l_{ds}|u)\!=\!p_{\Tb^{l_{sc}}, \lambdab^{l_{sc}}}(l_{sc}|u)  \cdot p_{\Tb^{l_{cc}}, \lambdab^{l_{cc}}}(l_{cc}|u) \cdot p_{\Tb^{l_{ds}}, \lambdab^{l_{ds}}}(l_{ds}|u)$} \nonumber
\end{align}
\begin{align}
   = &\prod_{i = 1} ^{d_{{sc}}}p_{\Tb^{l_{sc}}, \lambdab^{l_{sc}}}({l_{{sc}_i}}|u) \cdot \prod_{i = 1} ^{d_{{cc}}}p_{\Tb^{l_{cc}}, \lambdab^{l_{cc}}}({l_{{cc}_i}}|u) \cdot \prod_{i = 1}  ^{d_{{ds}}}p_{\Tb^{l_{ds}}, \lambdab^{l_{ds}}}({l_{{ds}_i}}|u) \nonumber \\
   \label{eq:expf3} 
   = &\prod_{Pa \in \{l_{sc}, l_{cc}, l_{ds} \}} \prod_{i = 1} ^{d_{Pa}}\frac{Q^{Pa}_i({Pa_i})}{Z^{Pa}_i(u)}\cdot \exp\left[\sum_{j=1}^{k_{Pa}} T^{Pa}_{i,j}({Pa_i}) \lambda^{Pa}_{i,j}(u)\right],
\end{align}
where ${\Phib}_x$ contains sufficient statistics $\Tb$ and coefficient $\lambdab$, $Q^{Pa}_i$ is the base measure, and $Z^{Pa}_i$ is the normalization function.

For $p_{f_x}(x|l_{sc}, l_{cc}, l_{ds})$, we constrain it by Additive Noise Model (ANM) assumption \cite{JanzingPMS09}, where the DGP for $x$ can be expressed as:
\begin{align}
    \label{eq:anm1}
    x &= f_x(l_{sc}, l_{cc}, l_{ds}) + \epsb, \epsb \sim p_\epsb(\epsb).
\end{align}

We rewrite Equation~\ref{eq:gen} using Equation~\ref{eq:anm1}, i.e.,
\begin{align}
    \label{eq:anm2}
    p_{\Theta_x}(x, Pa(x)|u) & = p_{\epsb}(x - f_x(l_{sc}, l_{cc}, l_{ds}))\cdot p_{{\Phib}_x}(l_{sc}, l_{cc}, l_{ds}|u).
\end{align}

Similarly, we can obtain the results for $y$, $o_{st}$ and $o_{ct}$, i.e.,
\begin{align}
     \label{eq:anm_y}
    p_{\Theta_y}(y, Pa(y)|u)  = p_{\epsb}(y - f_y(l_{ss}, l_{cc}))\cdot p_{{\Phib}_y}(l_{ss}, l_{cc}|u), \\ 
     \label{eq:anm_ts}
    p_{\Theta_{o_{st}}}(o_{st}, Pa(o_{st})|u)  = p_{\epsb}(o_{st} - f_{o_{st}}(l_{sc}))\cdot p_{{\Phib}_{o_{st}}}(l_{sc}|u),\\ 
     \label{eq:anm_tc}
    p_{\Theta_{o_{ct}}}(o_{ct}, Pa(o_{ct})|u)  = p_{\epsb}(o_{ct} - f_{o_{ct}}(l_{cc}))\cdot p_{{\Phib}_{o_{ct}}}(l_{cc}|u).
\end{align}

In summary, using the DGP in our SCM, we can express the joint probability density functions as Equations~\ref{eq:anm2}-\ref{eq:anm_tc}.

Notice that latent variables cannot be directly obtained. Instead, we can only learn their representations by mimicking these distributions. This raises a crucial question: Can we learn representations for each latent factor without mixing information with others, while ensuring that the difference between the learned representations and the true representations remains within acceptable bounds of uncertainty? This refers to the identifiability of the latent variables.

How to ensure identifiability, i.e, how to solve the question, is presented in the subsequent section.

\vspace{-2mm}
\subsection{Identifiability Analysis}
\label{sec:idf_ana}

As discussed in Section~\ref{section:SCM}, we aim to learn representations for latent factors while ensuring their identifiability. To achieve this, we begin by defining an equivalence relation denoted as $\sim_P$. 

\begin{definition}[$\sim_P$ Equivalent]
Suppose $\Theta$ and $\tilde{\Theta}$ are two set of parameters for the SCM as defined in Section~\ref{section:SCM}. $\Theta$ and $\tilde{\Theta}$ are  called  $\sim_P$ equivalent if the following conditions are met:
\begin{align}
    \label{eqn:def1}
    p_\Theta({o_{st}}, {o_{ct}}, x, y) &= p_{\tilde{\Theta}}({o_{st}}, {o_{ct}}, x, y), \\
    \label{eqn:def2}
   \forall o, \forall l, \exists (\mathbf{A}^{l},\mathbf{c}^l), \ \mathrm{ s.t. } \ \Tb^{l} ([f_{o}]_{l}^{-1}(o)) &= \mathbf{A}^{l} \tilde{\Tb}^{l} ([\tilde{f}_{o}]_{l}^{-1}(o)) + \mathbf{v}^{l}
\end{align}
where $o \in \{x, y, o_{ct}, o_{st}\}$ , $l$ is a latent factor in $Pa(o)$, expressed as $l \in Pa(o)$, $\Ab$ is an invertible permutation matrix, and $\mathbf{v}$ is a vector. 

\label{def:equivalence}
\end{definition}  

The following Theorem~\ref{thm:mainthm} provides a sufficient condition which ensures our model to learn parameters $\tilde{\Theta}$ that satisfy $\sim_p$ equivalence with true parameters $\Theta$.

\begin{figure*}[t]
  \centering
   \includegraphics[scale=0.6]{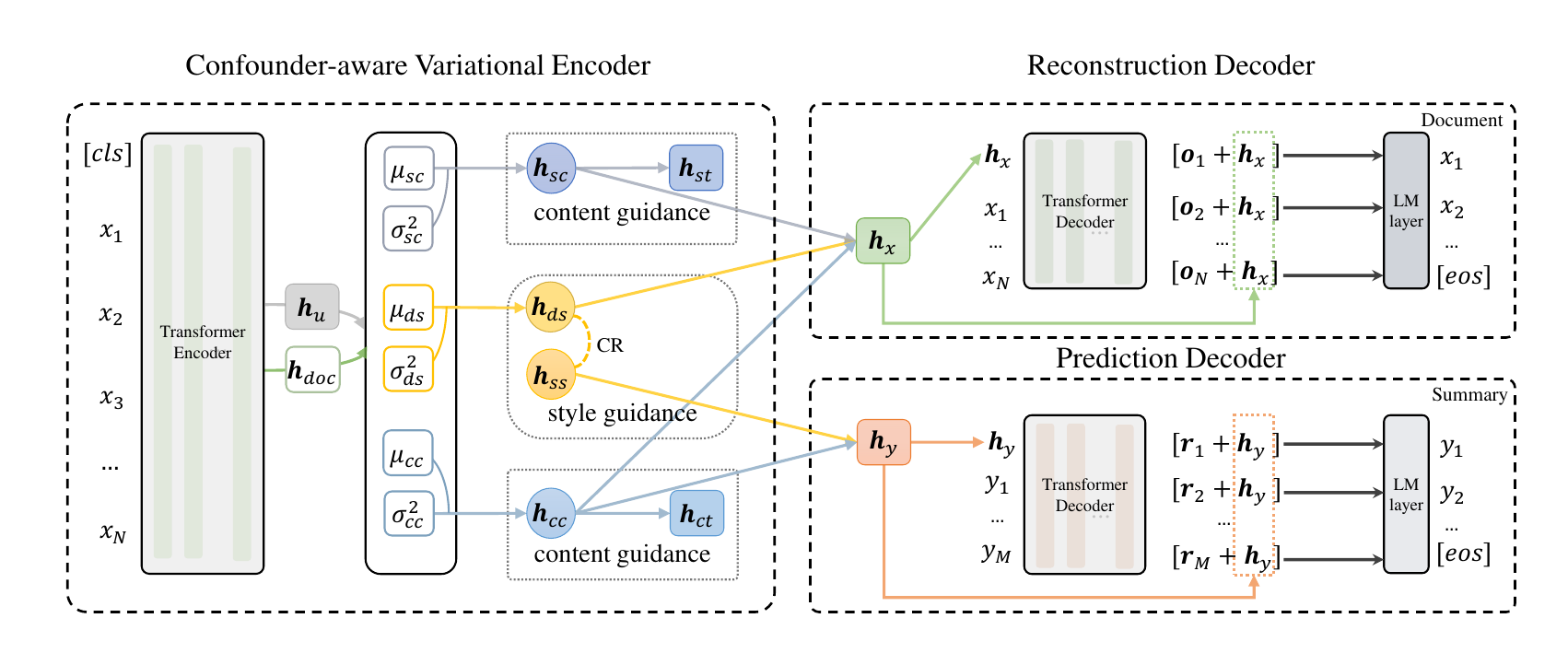}
 \vspace{-2mm}
   \caption{The overall architecture of CI-Seq2Seq model.}
   \label{fig:model}
   \vspace{-2mm}
\end{figure*}

\begin{theorem}[$\sim_P$ Identifiability]
\label{thm:mainthm}
Considering the SCM described in Section~\ref{section:SCM}, if we have an adequate number of distinct $u$ values, denoted as $k_{u}$, that satisfy the variety assumption, i.e., the matrix $\mathbf{L} \triangleq [\lambdab(u_1) - \lambdab(u_0),..., \lambdab(u_{k_u}) - \lambdab(u_0)]$ has full column rank, where $\lambdab(u)$ represents the vector parameter for the probability density function of an exponential family distribution. Then the learned parameter $\tilde{\Theta}$ and the true parameter $\Theta$ exhibit $\sim_p$ equivalent. 
\end{theorem}

\noindent\textbf{Discussion} Theorem~\ref{thm:mainthm} ensures that the learned parameters are $\sim_p$ equivalent with the true parameters, that is:
\begin{enumerate*}
    \item The joint distributions given by learned parameters and true parameters match (Equation~\ref{eqn:def1}).
    \item Latent factors can be separated, as only one appears in Equation~\ref{eqn:def2} each time.
    \item The difference between learned latent factors and true ones is limited to a permutation transformation with a linear shift applied to their sufficient statistics (Equation~\ref{eqn:def2}).
\end{enumerate*}

Besides, Theorem~\ref{thm:mainthm} requires that the number $k_u$ of different values for confounder $u$ is sufficient. 
It can be satisfied by proper definition for confounder.
Proofs are provided in Section~\ref{sec:proof}.

\section{Causality Inspired Seq2Seq Model}
\label{model overview}

Under the theoretical guarantees on modeling latent factors separately (Theorem~\ref{thm:mainthm}), we propose the Causality Inspired Sequence-to-Sequence (CI-Seq2Seq) model to learn representations that can mimic the latent factors by fitting the observed training data.

In the following, we first present our model architecture, a restructured Variational Auto-Encoder (VAE) \cite{kingma2013auto}, which learns latent representations from input and produces samples resembling the original data. 
Then, we detail learning strategy for causal generative mechanisms $p(x|l_{cc},l_{sc},l_{ds})$ and $p(y|l_{cc},l_{ss})$, followed by the controlled generation procedure using these learned mechanisms.

\subsection{Model Architecture}
As depicted in Figure \ref{fig:model}, the proposed CI-Seq2Seq contains three main components:  
Confounder-aware Variational Encoder, Reconstruction Decoder, and Prediction Decoder.

\vspace{-2mm}
\subsubsection{Confounder-aware Variational Encoder.}
This encoder targets to obtain representations $\textbf{h}_{cc}$, $\textbf{h}_{sc}$, $\textbf{h}_{ds}$ and $\textbf{h}_{ss}$ for the CC, SC, DS and SS factors from the input document $x=\{x_1,x_2,...,x_N\}$ of length $N$. 
Based on Theorem~\ref{thm:mainthm}, confounder $u$ is essential in distinguishing latent factors. It can be defined as the intrinsic properties of training data, e.g., topic, style and domains.
Here, we denote $u$ as the topic extracted from documents. 
Then, the encoder maps $x$ and $u$ into $\textbf{h}_{cc}$, $\textbf{h}_{sc}$, $\textbf{h}_{ds}$ and $\textbf{h}_{ss}$ according to $q_\phi (l_{cc},l_{sc},l_{ds}|x,u)$ and the relation between $l_{ds}$ and $l_{ss}$, where $\phi$ denotes the parameters of the confounder-aware variational encoder.

\begin{itemize}[leftmargin=*,itemsep=0pt,topsep=0pt,parsep=0pt]
    \item \textbf{Encoding Confounder Information $\textbf{h}_{u}$.}
    To achieve different values of confounder $u$, we denote $u$ as the topical features.
    We first partition each summarization corpus into $k_u$ subsets via LDA topic classification, where each document belongs to one subset.
    Specifically, each document obtains a topic distribution from LDA, and the topic id $tid$ with the highest probability is assigned to the document.    
    Then, following the practice of word embedding \cite{mikolov2013distributed}, $tid$ is applied to look up a hidden vector $\textbf{h}_{u}\in \mathbb{R}^{d_u}$ from a trainable embedding matrix $\textbf{E}_{u} \in \mathbb{R}^{k_u \times d_u}$, i.e., $ \textbf{h}_{u} = \textbf{E}_{u}(tid)$. 

    \item \textbf{Encoding Source Information $\textbf{h}_{doc}$.}
    CC, SC and DS factors are probably influenced by the full information of the document.
    Therefore, we propose to model the distribution of them conditioned on the global semantic representation of $x$. 
    Specifically, given an input document $x$, we first add a special token ``[CLS]'' in front of it, and then leverage the final hidden state of this token as its global representation $\textbf{h}_{doc}\in \mathbb{R}^{d_h}$. 
    It is a flexible aggregate and comprehensive understanding of the entire sequence.

    \item \textbf{Sampling $\textbf{h}_{cc}$, $\textbf{h}_{sc}$ and $\textbf{h}_{ds}$.}
    We mix $\textbf{h}_{doc}$ with $\textbf{h}_{u}$ and encode them into the distribution of $\textbf{h}_{cc}$, $\textbf{h}_{sc}$ and $\textbf{h}_{ds}$ to model the posterior distributions $p(l_{cc},l_{sc},l_{ds}|x,u)$.  
    Specifically, the true posterior $p(l_{cc},l_{sc},l_{ds}|x,u)$ is approximated via the variational distribution $q_\phi (l_{cc},l_{sc},l_{ds}|x,u)$. 
    We constraint the prior distributions $p(l_{cc},l_{sc},l_{ds})$ as standard Gaussian distributions following \cite{kingma2013auto,li2020optimus}. 
    Gaussian parameters mean $\mu_{cc,sc,ds}$ and variance $\sigma _{cc,sc,ds}^2$ are projected from the concatenation of $\textbf{h}_{doc}$ and $\textbf{h}_{u}$:
    \begin{gather}
        [\mu_{cc};\mu _{sc};\mu _{ds}] = \textbf{W}_{1} [\textbf{h}_{doc};\textbf{h}_{u}]+\textbf{b}_{1}, \\
        ([\log \sigma_{cc}^2;\log \sigma_{sc}^2;\log \sigma_{ds}^2])\! =\textbf{W}_{2} [\textbf{h}_{doc};\textbf{h}_{u}]+\textbf{b}_{2},
        \notag
    \end{gather}
    where $\textbf{W}_{1}$, $\textbf{W}_{2} \in \mathbb{R}^{(d_h+d_u) \times (d_{cc}+d_{sc}+d_{ds})}$, $\textbf{b}_{1}$, $\textbf{b}_{2} \in \mathbb{R}^{d_{cc}+d_{sc}+d_{ds}}$ are learnable.
    Finally, we sample $\textbf{h}_{cc}\in \mathbb{R}^{d_{cc}}$, $\textbf{h}_{sc}\in \mathbb{R}^{d_{sc}}$ and $\textbf{h}_{ds}\in \mathbb{R}^{d_{ds}}$ from the learned distribution using a reparametrization trick \cite{kingma2013auto}, respectively.

    \item \textbf{Computing $\textbf{h}_{ss}$ with Style Guidance.}
    Since the SS factors are only causally related to the summary $y =\{y_1,y_2,...,y_M\}$ of length $M$, it is not suitable to directly extract them from $x$ like DS factors. 
    Therefore, we introduce compression rate (CR) \cite{hahn2000challenges} between DS and SS factors as the style guidance. 
    Specifically, CR help bridge DS and SS factors smoothly, which indicates the information ratio between the target summary and the source document. Following previous work \cite{yeh2005text}, we define CR as the ratio of the text length between the summary and the document, i.e., $CR = M / N \in (0,1)$.
    Based on $CR$, we can obtain $\textbf{h}_{ss}$:
    \begin{equation}
        \textbf{h}_{ss} = \textbf{h}_{ds} \times CR
    \end{equation}
\end{itemize}

\subsubsection{Reconstruction Decoder.}
This decoder targets to utilize the representations $\textbf{h}_{cc}$, $\textbf{h}_{sc}$ and $\textbf{h}_{ds}$ of the CC, SC and DS factors to reconstruct the input document $x$ according to $p_{\theta}(x|l_{cc},l_{sc},l_{ds})$, where $\theta$ denotes the parameters of the reconstruction decoder.

First, we apply a fully connected (FC) layer to combine $\textbf{h}_{cc}$, $\textbf{h}_{sc}$ and $\textbf{h}_{ds}$ into the composed information $\textbf{h}_{x}$.
Then, we propose to replace the first token of decoder input with $\textbf{h}_{x}$, since the first token matters much for the generation of following tokens. 
Besides, the first token is only allowed to attend to itself, which could alleviate the vanishing latent factor problem to some extent \cite{xia2020composed}.   

To further enhance the impact of $\textbf{h}_{x}$, we add it to all the output hidden states $\{\textbf{o}_i\}_{i=1}^{N}$ from the last Transformer layer in the reconstruction decoder. The vocabulary selection probability $P_x$ for generating $x$ is computed as 
\begin{equation}
  P_x = \textbf{W}_{3}(\textbf{o}_i+\textbf{h}_{x})+\textbf{b}_{3},
\end{equation}
where $\textbf{W}_{3} \in \mathbb{R}^{d_h\times d_v}$ and $\textbf{b}_{3} \in \mathbb{R}^{d_v}$ are learnable.

\subsubsection{Prediction Decoder.}
This decoder only allows the injection of the CC representation $\textbf{h}_{cc}$ along with the SS representation $\textbf{h}_{ss}$ for generating the summary $y$ according to $p_{\tau}(y|l_{cc},l_{ss})$, where $\tau$ denotes the parameters of the prediction decoder.

First, similar to the reconstruction decoder, we obtain the composed representation $\textbf{h}_y$ for summary prediction, by combining $\textbf{h}_{cc}$ and $\textbf{h}_{ss}$ using a FC layer.
Then, we replace the first token with $\textbf{h}_y$ in the prediction decoder.  
Simultaneously, we add $\textbf{h}_y$ to all the output hidden states $\{\textbf{r}_j\}_{j=1}^{M}$ from the last transformer layer in the prediction decoder.
The final vocabulary selection probability $P_y$ for generating $y$ is calculated in the same way as $P_x$.

\subsection{Learning Strategy}
To learn $p(x|l_{cc},l_{sc},l_{ds})$ and $p(y|l_{cc},l_{ss})$ for invariant prediction, we reformulate the learning objective function of VAE in the supervised scenario to fit the training corpus. 
Specifically, we apply four learning objectives as follows.

\begin{itemize}[leftmargin=*,itemsep=0pt,topsep=0pt,parsep=0pt]
    \item \textbf{Reconstruction Loss} is applied to train the reconstruction decoder to reconstruct the input document, i.e., 
    \begin{equation}
      \mathcal{L}_{R}=-\mathbb{E}_{q_\phi(l_{cc},l_{sc},l_{ds}|x,u)}[\log p_\theta(x|l_{cc},l_{sc},l_{ds})].
    \end{equation} 
    
    \item \textbf{Prediction Loss} is applied to encourage the prediction decoder to generate the summary based on the summary-causal representations, i.e.,
    \begin{equation}
      \mathcal{L}_{P}=-\mathbb{E}_{q_\phi(l_{cc},l_{sc},l_{ds}|x,u)}[\log p_\tau(y|l_{cc},l_{ss})].
    \end{equation}
    
    \item \textbf{KL Loss} is a regularizer based on the Kullback-Leibler (KL) divergence, applied to push the posterior $q_\phi(l_{cc},l_{sc},l_{ds}|x,u)$ to be closed to the prior $p(l_{cc},l_{sc},l_{ds})$ which is constrained as standard Gaussian distributions, i.e., 
    \begin{equation}
        \mathcal{L}_{KL}=\mathbb{D}_{KL}[q_\phi(l_{cc},l_{sc},l_{ds}|x,u) \Vert p(l_{cc},l_{sc},l_{ds})].
    \end{equation}
    \item \textbf{Content Guidance Loss} is further applied to guide the optimization of the CC and SC factors, which is calculated by three steps.
    (i) We first extract the core topics $o_{ct}$ and side topics $o_{st}$ in $x$ according to the LDA topic distribution $p(o_t|x)$ on $k_u$ topics.
    Specifically, given a threshold $th$, a topic $o_t^a (a\in\{1,2,...,k_u\})$ belongs to the core topics of document $x$ if $p(o_t=o_t^a|x) > th$, otherwise side ones.
    To indicate the type of each topic, we introduce a $k_u-$dimension binary indicator $\textbf{g}$, where ‘‘1'' represents the core topics and ‘‘0'' represents the side ones.
    (ii) We then transform such topic information into hidden representations $\textbf{h}_{ct}, \textbf{h}_{st}\in \mathbb{R}^{d_h}$ based on another learnable embedding matrix $\textbf{E}_{t}\in \mathbb{R}^{{k_u} \times d_h}$.
    Similar to $\textbf{E}_u$, each row of $\textbf{E}_{t}$ represents a topic embedding.
    Specifically, to achieve the aggregated hidden representation $\textbf{h}_{ct}$ which combines information of all core topics, we obtain the core topic distribution $p(o_{ct}|x)$ based on the binary indicator $\textbf{g}$, i.e.,
    \begin{equation}
        p(o_{ct}|x) = Norm(p(o_t|x)\odot \textbf{g}),
    \end{equation}
    where $\odot$ denotes element-wise multiplication and $Norm()$ denotes normalization operation.
    After that, we linearly combine topic embeddings from $\textbf{E}_{t}$ according to $p(o_{ct}|x)$ as below:
    \begin{equation}
        \textbf{h}_{ct} = p(o_{ct}|x)\textbf{E}_{t}.
    \end{equation}
    Similarly, for side topics, we have
    \begin{gather}
        p(o_{st}|x) = Norm(p(t|x)\odot (\textbf{1}-\textbf{g})), \\
        \textbf{h}_{st} = p(o_{st}|x)\textbf{E}_{t}.
    \end{gather}
    (iii) Finally, we compute the Euclidean distance (i.e., L2 distance) of $\langle \textbf{h}_{cc},\textbf{h}_{ct}\rangle$ and $\langle \textbf{h}_{sc},\textbf{h}_{st}\rangle$ as the content guidance loss, i.e.,
    \begin{equation}
        \label{eqn:lda-loss}
        \mathcal{L}_{LDA}= L2(\textbf{h}_{cc}, \textbf{h}_{ct}) + L2(\textbf{h}_{sc}, \textbf{h}_{st}).
    \end{equation}
    
\end{itemize}

The total loss is a summation of the four losses:
\begin{equation}
  \mathcal{L} = \mathcal{L}_{R} + \mathcal{L}_{P} + \lambda_{kl} \mathcal{L}_{KL} +  \lambda_{lda} \mathcal{L}_{LDA},
\end{equation}
where $\lambda_{kl}$ and $\lambda_{lda}$ are used to control the strength of the regularization and the content guidance.

\subsection{Controlled Generation}
During the test stage, we first optimize the following log-likelihood to infer $l_{cc}^{*}$, $l_{sc}^{*}$ and $l_{ds}^{*}$, i.e.,
\begin{equation}
\begin{aligned}
\label{eqn:infer}
    \max_{l_{cc},l_{sc},l_{ds}} &\log p_{\theta}(x|l_{cc},l_{sc},l_{ds})\!+\! \lambda_{cc} \Vert l_{cc}\Vert_2^2 \!+\! \lambda_{sc} \Vert l_{sc}\Vert_2^2 \!+\! \lambda_{ds} \Vert l_{ds}\Vert_2^2,
\end{aligned}
\end{equation}
where $\lambda_{cc}$, $\lambda_{sc}$ and $\lambda_{ds}$ control the learned $l_{cc}$, $l_{sc}$ and $l_{ds}$ in a reasonable scale. 
Specifically, we sample some candidate points from $N(0,I)$ and select the optimal one in terms of Equation \ref{eqn:infer} as the initial point for further optimization.

Finally, we employ the optimized $l_{cc}^{*}$ and $l_{ds}^{*}$ to generate summaries with different styles by varying $CR$\footnote{Note that $CR$ is the ground-truth summary-document length ratio during training.}. 
In this way, we can actively control the compression rate of the summary. 
That is, with different $l_{ss}^{*}$ values and the optimized $l_{cc}^{*}$, we generate the summary $y$ based on the learned $p_\tau(y|l_{cc}^{*},l_{ss}^{*})$.

\section{Proof}

\begin{proof}
\label{sec:proof}
For Theorem~\ref{thm:mainthm}, we will demonstrate that we can learn a parameter $\tilde{\Theta}$ that is $\sim_P$ equivalent to the true parameter $\Theta$, satisfying two conditions: Equation~\ref{eqn:def1} and Equation~\ref{eqn:def2} in Definition~\ref{def:equivalence}. The first condition means the correct fitting of the joint distribution of observed variables, which can be guaranteed by the universal approximation ability of neural networks. Therefore, our main task is to prove the validity of the second condition. 

The proof is roughly divided into two steps: \emph{Denoising} and \emph{Identifying}. We will present the proof step by step.

\noindent \textbf{(1) Denoising.}
This step serves the purpose of eliminating noise variables while retaining only the latent factors. Assuming that the learned distribution of observed variables is identical to the true distribution (i.e., the first condition holds), we have $p_\Theta({o_{st}}, {o_{ct}}, x, y) = p_{\tilde{\Theta}}({o_{st}}, {o_{ct}}, x, y)$. By integrating variables $o_{st}$, $o_{ct}$, and $y$, we can obtain $p_{\Theta_{x}}(x) = p_{\tilde{\Theta}_{x}}(x)$. 
Then, we express it given confounder:
\begin{equation}
    \label{eqn:befinsert}
    \resizebox{1\hsize}{!}{$p_{f_{x}, \Tb^{l_{sc}}, \lambdab^{l_{sc}}, \Tb^{l_{ds}}, \lambdab^{l_{ds}}, \Tb^{l_{cc}}, \lambdab^{l_{cc}}}(x|u)
    \!= \!p_{\tilde{f}_{x}, \tilde{\Tb}^{l_{sc}}, \tilde{\lambdab}^{l_{sc}}, \tilde{\Tb}^{l_{ds}}, \tilde{\lambdab}^{l_{ds}}, \tilde{\Tb}^{l_{cc}}, \tilde{\lambdab}^{l_{cc}}}(x|u),$}
\end{equation}
\begin{flalign}
   \Rightarrow \int &\resizebox{0.793\hsize}{!}{$p_{\epsb}(x - f_x(l_{sc}, l_{cc}, l_{ds})) p_{{\Phib}_x}(l_{sc}, l_{cc}, l_{ds}|u) \cdot dl_{sc} dl_{cc} dl_{ds}$} & \nonumber \\
   = \int &\resizebox{0.793\hsize}{!}{$p_{\epsb}(x - \tilde{f}_x(l_{sc}, l_{cc}, l_{ds})) p_{\tilde{\Phib}_x}(l_{sc}, l_{cc}, l_{ds}|u) \cdot dl_{sc} dl_{cc} dl_{ds},$} &
\end{flalign}
\begin{flalign}
   \label{eqn:jacobian}
   \Rightarrow \int &p_{\epsb}(x - \bar{x}) p_{{\Phib}_x}(f_x^{-1}(\bar{x})|u) \left|\det(J_{f_x^{-1}}(\bar{x}))\right| d\bar{x} & \nonumber \\
   = \int &p_{\epsb}(x - \tilde{x}) p_{\tilde{\Phib}_x}(\tilde{f}_x^{-1}(\tilde{x})|u) \left|\det(J_{\tilde{f}_x^{-1}}(\tilde{x})\right| d\tilde{x}, &
\end{flalign}
\begin{flalign}
    \label{eqn:newfunc}
   \Rightarrow \int &\resizebox{0.793\hsize}{!}{$p_{\epsb}(x - \bar{x}) p_{{\Phib}_x, f_x, u}(\bar{x})d\bar{x} = \int p_{\epsb}(x - \tilde{x}) p_{\tilde{\Phib}_x, \tilde{f}_x, u}(\tilde{x})d\tilde{x},$} & 
\end{flalign}
\begin{flalign}
    \label{eqn:conv0}
    \Rightarrow \quad &(p_{\epsb}* p_{{\Phib}_x, f_x, u})(x) = (p_{\epsb} * p_{\tilde{\Phib}_x, \tilde{f}_x, u})(x), & 
\end{flalign}
\begin{flalign}
   \label{eqn:conv}
   \Rightarrow \quad &F[p_{\epsb}](\omega)F[p_{{\Phib}_x, f_x, u}](\omega) =F[p_{\epsb}](\omega)F[p_{\tilde{{\Phib}}_x, \tilde{f}_x, u}](\omega), &
\end{flalign}
\begin{flalign}
   \label{eqn:xres}
   \Rightarrow \quad &p_{{\Phib}_x, f_x, u}(x)=p_{\tilde{{\Phib}}_x, \tilde{f}_x, u}(x). &
\end{flalign}

In Equation~\ref{eqn:jacobian}, $J$ represents the Jacobian matrix, and $\left|\det\right|$ denotes generalized determinant, defined as $\left|\det (A) \right| \triangleq \sqrt{\det(A^\top A)}$. The symbol $\triangleq$ is read as "is defined as". Equation~\ref{eqn:newfunc} introduces the function $p_{{\Phib}x, f, u}(\bar{x}) \triangleq p_{{\Phib}x}(f_x^{-1}(\bar{x})|u)$ $\det(J_{f_x^{-1}}(\bar{x}))$ for convenience. Note that Equation~\ref{eqn:newfunc} corresponds to a convolution operation as expressed in Equation~\ref{eqn:conv0}. In Equation~\ref{eqn:conv}, $F$ means Fourier transformation, a useful tool to simplify convolution. In Equation~\ref{eqn:xres}, we obtain the denoised result. Similar results can be obtained for the other observed variables $y$, $o_{st}$, and $o_{ct}$, i.e.,
\begin{align}
    \label{eqn:yres}
    p_{\Phib_y, f_y, u}(y)&=p_{\tilde{\Phib}_{y}, \tilde{f}_y, u}(y), \\
    \label{eqn:tsres}
    p_{\Phib_{o_{st}}, f_{o_{st}}, u}(o_{st})&=p_{\tilde{\Phib}_{o_{st}}, \tilde{f}_{o_{st}}, u}(o_{st}), \\
    \label{eqn:tcres}
    p_{\Phib_{o_{ct}}, f_{o_{ct}}, u}(o_{ct})&=p_{\tilde{\Phib}_{o_{ct}}, \tilde{f}_{o_{ct}}, u}(o_{ct}).
\end{align}

Furthermore, notice that different observed variables share common latent factors. To capture this characteristic, we specifically target pairs of observed variables and apply the aforementioned denoising method to these pairs. This idea is inspired by LaCIM \cite{sun2021recovering}. For the variable pairs ($x$, $o_{st}$), ($x$, $o_{ct}$) and ($y$, $o_{ct}$), we can obtain the similar denoised results:
\begin{align}
    \label{eqn:ares}
    p_{{\Phib}_x, \Phib_{o_{st}}, f_a, u}(a)&=p_{\tilde{\Phib}_x, \tilde{\Phib}_{o_{st}}, \tilde{f}_a, u}(a), \\
    \label{eqn:bres}
    p_{{\Phib}_x, \Phib_{o_{ct}}, f_b, u}(b)&=p_{\tilde{\Phib}_x, \tilde{\Phib}_{o_{ct}}, \tilde{f}_b, u}(b), \\
    \label{eqn:cres}
    p_{\Phib_y, \Phib_{o_{ct}}, f_c, u}(c)&=p_{\tilde{\Phib}_y, \tilde{\Phib}_{o_{ct}}, \tilde{f}_c, u}(c),
\end{align}
where $a \triangleq [x^\top, o_{st}^\top]^\top$, $f_a^{-1} \triangleq [[f_x]^{-1}_{l_{ds}, l_{cc}}(x)^\top,f_{o_{st}}^{-1}(o_{st})^\top]^\top$, $b \triangleq [x^\top, o_{ct}^\top]^\top$, $f_b^{-1} \triangleq [[f_x]^{-1}_{l_{sc}, l_{ds}}(x)^\top,$ $f_{o_{ct}}^{-1}(o_{ct})^\top]^\top$, $c \triangleq [y^\top, o_{ct}^\top]^\top$, $f_c^{-1} \triangleq [[f_y]^{-1}_{s_c}(y)^\top$, $f_{o_{ct}}^{-1}(o_{ct})^\top]^\top$.

\noindent \textbf{(2) Identifying.}
This step aims to establish the validity of Equation~\ref{eqn:def2}, which asserts the identifiability of each latent factors. Firstly, we present the process for separating these variables.  Subsequently, we will transform the resulting equations to derive Equation~\ref{eqn:def2}.

Considering that we have sufficient number $k_u$ of different values of $u$. Taking the logarithm on the both sides of Equations~\ref{eqn:xres}-\ref{eqn:cres}, then we plug these different $u$ (i.e., $u_0, u_1, ...u_{k_u}$) into each equation. Subtracting
the first equation (containing $u_0$) from the second equation ($u_1$) to the last equation ($u_{k_u}$), we obtain $k_u$ different equations for each of Equations~\ref{eqn:xres}-\ref{eqn:cres}, indexing by $q = 1, 2, \dots, k_u$:

 \begin{align}
    \label{eqn:midts}  
    & \sum_{l\in Pa(o)} \left[\dotprod{\Tb^{l}(f_{o}^{-1}(o))}{\overline{\lambdab^{l}}(u_q)} + \sum_i \log \frac{Z_i^{l}(u_0)}{Z_i^{l}(u_q)}\right]\ \nonumber \\
    = & \sum_{l \in Pa(o)} \left[\dotprod{\tilde{\Tb}^{l}(\tilde{f}_{o}^{-1}(o))}{\overline{\tilde{\lambdab^{l}}}(u_q)} + \sum_i \log \frac{\tilde{Z}^{l}_i(u_0)}{\tilde{Z}^{l}_i(u_q)}\right] .
 \end{align}
 In Equation~\ref{eqn:midts}, $o$ represents both observed variables and the variable pairs, i.e., $o \in \{x, y, o_{st}, o_{ct}, a, b, c\}$, where $a, b, c$ are the variable pairs defined in the first step. When $o$ represents variable pair, such as $a = [x^\top, o_{st}^\top]$, then $Pa(o = a) = Pa(x) \cup Pa(o_{st})$. And we define $\overline{\lambdab^{l}}(u_q) \triangleq \lambdab^{l}(u_q) - \lambdab^{l}(u_0)$. In order to further simplify Equation~\ref{eqn:midts}, we define $\wb^{l}_q \triangleq \sum_{i}\frac{\tilde{Z}^{l}_i(u_0)Z_i^{l}(u_q)}{\tilde{Z}^{l}_i(u_q)Z_i^{l}(u_0)}$. Then we rewrite these equations in matrix form:
 \begin{align}
    \label{eqn:osep}
    \sum_{l\in Pa(o)} \left[L^{l, \top} \Tb^{l}(f_{o}^{-1} (o))\right] = \sum_{l\in Pa(o)} \left[\tilde{L}^{l, \top} \tilde{\Tb}^{l}(\tilde{f}_{o}^{-1}(o)) + \wb^{l}\right].
 \end{align}

We denote Equation~\ref{eqn:osep} as Eq($\cdot$). Notice that $Pa(x) = \{l_{sc}, l_{ds}, l_{cc}\}$, we will now outline the procedure for separating the latent factors in the parent set of $x$. By evaluating the expression $Eq(o = x) + Eq(o = o_{st}) - Eq(o = a)$, we can separate the latent factor $l_{sc}$ of observed variable $x$:
\begin{align}
    \label{eqn:xcs}
    L^{l_{sc}, \top} \Tb^{l_{sc}}([f_{x}]_{l_{sc}}^{-1} (x)) = \tilde{L}^{l_{sc}, \top} \tilde{\Tb}^{l_{sc}}([\tilde{f}_{x}]_{l_{sc}}^{-1} (x)) + \wb^{l_{sc}}.
\end{align}
Using the same method we can separate $l_{cc}$ of $x$ by evaluating $Eq(o = x) + Eq(o = o_{ct}) - Eq(o = b)$:
\begin{align}
    \label{eqn:xcc}
    L^{l_{cc}, \top} \Tb^{l_{cc}}([f_{x}]_{l_{cc}}^{-1} (x)) = \tilde{L}^{l_{cc}, \top} \tilde{\Tb}^{l_{cc}}([\tilde{f}_{x}]_{l_{cc}}^{-1} (x)) + \wb^{l_{cc}}.
\end{align}
Afterwards, the only remaining latent factor of $x$, $l_{ds}$, is naturally separated. The above results show that for the observed variable $x$, all of its latent factors can be separated while preserving their individuality, without mixed information. This conclusion also holds true for other observed variables, i.e., $y$, $o_{ct}$, and $o_{st}$. The equations for each separated latent factors can be expressed as follows:
\begin{equation}
    \label{eqn:all_sep}
    L^{l, \top} \Tb^{l}([f_{o}]_{l}^{-1} (o)) = \tilde{L}^{l, \top} \tilde{\Tb}^{l}([\tilde{f}_{o}]_{l}^{-1} (o)) + \wb^{l}, 
\end{equation}
where $o \in \{x, y, o_{st}, o_{ct}, a, b, c\}$, $l \in Pa(o)$.

Based on Equation~\ref{eqn:all_sep}, we will demonstrate the validity of Equation~\ref{eqn:def2}. Since number $k_u$ is enough to ensure matrix $L^{l, \top}$ has full rank, we multiply it's inverse matrix on both sides of Equation~\ref{eqn:all_sep}:
\begin{align}
    \label{eqn:sep_final}
    \Tb^{l}([f_{o}]_{l}^{-1} (o)) = \Ab^{l} \tilde{\Tb}^{l}([\tilde{f}_{o}]_{l}^{-1} (o)) + \vb^{l},
\end{align}
where $\Ab^{l} = (L^{l, \top})^{-1}\tilde{L}^{l, \top}$, $\vb^{l} = (L^{l, \top})^{-1}$ $\wb^{l}$. Notice that Equation~\ref{eqn:sep_final} is already in the same form as Equation~\ref{eqn:def2}. 

The remaining task is to prove that the matrix $\Ab^l$ is an invertible permutation matrix, which can be achieved by directly applying Lemma 3, Theorem 2, and Theorem 3 from \cite{khemakhem2020variational}.
\end{proof}

\vspace{-3mm}
\section{Experimental Settings}
\subsection{Datasets}

We conduct experiments on two public text summarization datasets in English: 
(1) \textbf{XSUM} \cite{narayan2018don} contains BBC articles accompanied with single sentence summaries (training/validation/testing size are 204,045/11,332/11,334 respectively); and (2) \textbf{CNN/DM} \cite{hermann2015teaching} contains news articles from CNN and Daily Mail websites paired with multi-sentence human generated summaries (training/validation/testing size are 286,817/13,368/11,487 respectively).

\vspace{-2mm}
\subsection{Evaluation Methodology}

\begin{itemize}[leftmargin=*,itemsep=0pt,topsep=0pt,parsep=0pt]
    \item \textbf{Automatic Evaluation:}
    We adopt \textbf{Rouge scores} \cite{lin2004rouge} to automatically evaluate the quality of the summaries generated by our model and the baselines.
    Specifically, we use \emph{Rouge-1 (R1)}, \emph{Rouge-2 (R2)} and \emph{Rouge-L (RL)} to measure the the uni-gram, bi-gram and longest-common subsequence similarities, respectively.

    \item \textbf{Human Evaluation:} We measure the \textbf{Informativeness}, \textbf{Faithfulness}, and \textbf{Fluency} referring to \cite{fabbri2021summeval,kryscinski2019neural,huang2020knowledge}.
    Each summary is rated on a 5-point Likert scale (higher better), to measure whether the generated summary can satisfy:
    \begin{enumerate*}[label=(\roman*)]
        \item \emph{Informativeness}, covering core information (i.e., the most necessary pieces) of the source document and excluding side information that may mislead the understanding of the main idea of the document;
        \item \emph{Faithfulness}, containing only information present in the document, without introducing any made-up facts (i.e., hallucination \cite{xiao2021hallucination});
        \item \emph{Fluency}, being natural and grammatically correct.
    \end{enumerate*}
    Specifically, we ask three college students to score 200 samples randomly picked from the test set of CNN/DM and XSUM (100 for each).

\end{itemize}

\vspace{-2mm}
\subsection{Baselines}
We compare CI-Seq2Seq against several recently proposed baseline methods: 
\begin{enumerate*}[label=(\roman*)]
    \item \textbf{Unified VAE-PGN} \cite{choi-etal-2019-vae} leverages VAE to eliminate non-critical information at a sentence-level for abstractive summarization.
    \item \textbf{VHTM} \cite{fu2020document} jointly accomplishes summarization with topic inference via variational encoder-decoder.
    \item \textbf{T5} \cite{raffel2020exploring} is a pre-trained framework that converts all text-based language problems into a text-to-text format.
    \item \textbf{BART} \cite{lewis-etal-2020-bart} is a denoising autoencoder for pre-training Seq2Seq models.
    \item \textbf{GLM} \cite{du-etal-2022-glm} is a General Language Model pre-trained with autoregressive blank infilling.
    \item \textbf{PtLAAM} \cite{liu-etal-2022-length} uses a length-aware attention mechanism to generate summaries with desired length.
    \item \textbf{PEGASUS} \cite{zhang2020pegasus} is a pre-trained model tailored for abstractive summarization, with Gap Sentences Generation (GSG) as pre-training objective.
\end{enumerate*}

\vspace{-2mm}
\subsection{Implementation Details}
The proposed CI-Seq2Seq can be adapted to other Seq2Seq PLMs.
Here, we choose BART-large and PEGASUS-large for initialization, denoted as CI-Seq2Seq$^{bart}$ and CI-Seq2Seq$^{pega}$, where the hidden size $d_h$ is 1024, and the vocabulary size $d_v$ is 50265 or 96103 for CI-Seq2Seq$^{bart}$ and CI-Seq2Seq$^{pega}$, respectively.
BART is chosen for its outstanding performance as well as less computing cost than its peers \cite{liu-etal-2022-length}, and PEGASUS is chosen for its state-of-the-art performance in summarization.
The number of new parameters added to CI-Seq2Seq compared to backbones is about 256M.

For hyper parameters, we use grid search to automatically find the best setup based on the validation set.
We select $d_{u}$ as 16 from $[8,32]$, $k_{u}$ as 5 from $[1,20]$, and $th$ as 0.25 from $[0.02,0.3]$.
We choose $d_{ds}$ and $d_{ss}$ as 128 from $\{128,256\}$, $d_{sc}$ as 256 from $\{256,512\}$, and $d_{cc}$ as 128 from $\{128,256\}$.
Note that the dimension of the SC representations $d_{sc}$ is set larger than that of the CC representations $d_{cc}$, for it is very likely that the SC representations include more diverse information than the CC representations describing the core information.
During training, we select $\lambda_{kl}$ and $\lambda_{lda}$ as 1 from $[1e^{-3},1]$.
The batch size is searched from $\{256,512\}$, and the learning rate is searched from $[1e^{-5},1e^{-4}]$.
During test, we select the best number of candidate points in the range of $[5,20]$ and the best optimization steps in the range of $[20,100]$. 
$\lambda_{cc}, \lambda_{sc}, \lambda_{ds}$ are searched from $[0.001,0.5]$, batch size is searched from $[1,4]$, and learning rate is searched from $[0.001,0.5]$.

Adam optimizer is utilized at both stages.
We train our model on one NVIDIA Tesla V100 32GB GPUs for about 5k$\sim$10k steps for each dataset, which takes approximately six days.
All experimental results are reported on the test set.
Note that for baseline methods, we reproduce and evaluate our backbone models (i.e., BART and PEGASUS) ourselves to provide a fair comparison, while we report scores of other baselines from the papers.
For BART, the results reproduced ourselves are almost consistent with those of the original paper \cite{lewis-etal-2020-bart}.
For PEGASUS, the results on XSUM are almost consistent. 
However, our reproduced results and the reported results in the original paper \cite{zhang2020pegasus} have a gap\footnote{Our reproduced results were consistent with directly leveraging the checkpoint from https://huggingface.co/google/pegasus-cnn\_dailymail.}.
The result difference between this work and the original paper may come from our restriction on the maximum sequence length, which is set to 512 for the source documents and 64 for the summaries.

\begin{table*}[t]
  \renewcommand{\arraystretch}{1}
  \setlength\tabcolsep{15pt}
    \caption{In-domain performance comparisons between our CI-Seq2Seq and the baselines on XSUM and CNN/DM datasets. 
    Best results are marked in boldface. * indicates statistically significant improvements over baselines (p-value < 0.05).}
    \label{tab:Comparisons}
    \centering
    \begin{tabular}{lcccccccccc}
      \toprule
      \multirow{3}{*}{Method}   & \multicolumn{3}{c}{XSUM} & \multicolumn{3}{c}{CNN/DM}\\
  
      \cmidrule(r){2-4}
      \cmidrule(r){5-7}
       & Rouge-1 & Rouge-2  & Rouge-L & Rouge-1 & Rouge-2  & Rouge-L \\
      \midrule
     Unified VAE-PGN \cite{choi-etal-2019-vae} &- &- &- & 39.32 &17.07 &29.43 \\
     VHTM \cite{fu2020document} &- &- &- & 40.57 &18.05 &37.18  \\
     T5 \cite{raffel2020exploring} &- &- & - & 42.50 &20.68& 39.75\\
     BART &45.02 &21.65& 36.56 & 43.84 &20.95 &40.92 \\
     GLM \cite{du-etal-2022-glm} & 45.50 & 23.50 & 37.30 &43.80 &21.00 & 40.50 \\
     PtLAAM \cite{liu-etal-2022-length} &45.53 &21.82 & 36.85 &44.21 &20.77&40.97\\
     PEGASUS &47.11 & 24.32 & 38.98 & 42.23 & 20.01 & 38.92\\

      \midrule 
      CI-Seq2Seq$^{bart}$ & 48.17* & 25.41* &40.24* & \textbf{45.05}* & 22.01* & \textbf{41.96}* \\
      CI-Seq2Seq$^{pega}$ & \textbf{51.07}* & \textbf{28.68}* &\textbf{44.04}* & 44.48 & \textbf{22.88}* & 41.30  \\
      
      \bottomrule
    \end{tabular}
     \vspace{-3mm}
\end{table*}

\vspace{-1mm}
\section{Experimental Results}

We aim to answer four research questions: 
\textbf{(RQ1)} Does CI-Seq2Seq enhance prediction performance on in-domain datasets?
\textbf{(RQ2)} Does CI-Seq2Seq enhance generalization ability on out-of-domain datasets?
\textbf{(RQ3)} Is CI-Seq2Seq Interpretable?
\textbf{(RQ4)} How do latent factors and their constraints affect the performance of CI-Seq2Seq?
For each question, we conduct experiments as follow.

\vspace{-2mm}
\subsection{In-domain Prediction Performance}
To answer \textbf{RQ1}, we compare CI-Seq2Seq with various strong baselines on the test set of CNN/DM and XSUM, where models are trained on the training set of the same corpus. 

\vspace{-2mm}
\subsubsection{Automatic Evaluation}
We have the following observations from Table \ref{tab:Comparisons}:
\begin{enumerate*}[label=(\roman*)]
    \item  VAE-based neural summarization models (i.e., \textit{Unified VAE-PGN} and \textit{VHTM}) perform well by automatically learning text representations containing critical information of documents.
    \item  The improvements of PLMs (i.e., \textit{T5}, \textit{BART} and \textit{GLM}) over previous methods demonstrate the utility of pre-training on massive corpora for downstream summarization tasks.
    \item By incorporating length-aware attention mechanism, \textit{PtLAAM} could further enhance the performance of BART.
    \item \textit{PEGASUS} outperforms all baselines on XSUM, showing the power of its tailored pre-training objective for summarization.
    On CNN/DM, PEGASUS performs less well than models initialized with BART under the same maximum sequence length constraint.
    The reason may lie in the different matching degree between the pre-training objective and the downstream datasets. 
    Specifically, BART's denoising objective is to reconstruct full text, while the GSG objective for PEGASUS is to reconstruct corrupted text. Consequently, BART, with its longer target text, can better handle the long summaries of CNN/DM than PEGASUS.

\end{enumerate*}

When we look at our CI-Seq2Seq model, we can find that:
\begin{enumerate*}[label=(\roman*)]
    \item Our \textit{CI-Seq2Seq} implemented by both BART and PEGASUS can outperform all the baselines on two datasets. 
    For example, CI-Seq2Seq$^{pega}$ performs 12.98\% better than PEGASUS on XSUM in terms of RL. 
    This indicates the insufficiency of only modeling statistical dependence and the effectiveness of modeling the causal relationships between observed documents and summaries.
    \item Between them, CI-Seq2Seq$^{pega}$ performs better on XSUM, while CI-Seq2Seq$^{bart}$ performs better on CNN/DM.
    Under the same fine-tuning setting, the possible explanation aligns with that accounting for the performance difference between BART and PEGASUS.
\end{enumerate*}

\subsubsection{Human Evaluation}
As shown in Table \ref{tab:human_evaluation}, we can observe that:
\begin{enumerate*}[label=(\roman*)]
    \item \textbf{Informativeness}: CI-Seq2Seq models implemented by two backbones perform better than baselines.
    It indicates that introducing causality helps to extract the core information into summaries, meanwhile effectively reducing the interference of side information.
    This is consistent with our purpose of distinguishing core content from side one in the document and leveraging the causal part for summary generation.
    \item \textbf{Faithfulness}: CI-Seq2Seq models also outperform baselines, indicating that our method could alleviate the hallucination by pursuing only core information in the document, though the hallucination problem is not our focus and deserves further exploration.
    \item \textbf{Fluency}: CI-Seq2Seq models are comparable to baselines, indicating that our models can retain the ability to generate fluent text while removing non-essential information. 
\end{enumerate*}

\begin{table}[h]
 \vspace{-2mm}
  \renewcommand{\arraystretch}{1}
  \setlength\tabcolsep{4pt}
    \caption{Average scores of human evaluation about Informativeness (Info.), Faithfulness (Faith.) and Fluency (Flu.). The consistency between annotators is measured by Fleiss's kappa, which is 0.71.}
    \label{tab:human_evaluation}
    \centering
    \begin{tabular}{ccccccc}
      \toprule
  \multirow{2}{*}{Method} & \multicolumn{3}{c}{XSUM} & \multicolumn{3}{c}{CNN/DM} \\
      \cmidrule(r){2-4}
      \cmidrule(r){5-7}
      & Info. & Faith. & Flu. & Info. & Faith. & Flu. \\
      \midrule
      BART
      & 3.25 & 3.99 & 4.86 & 2.73 & 4.95 & 4.78 \\
      PEGASUS
      & 3.57 & 4.21 & 4.92 & 3.13 & 4.95 & 4.88 \\
      \hline
      CI-Seq2Seq$^{bart}$
      & 4.01 & 4.35 & 4.91 & \textbf{3.62} & 4.96 & \textbf{4.92} \\
      CI-Seq2Seq$^{pega}$
      & \textbf{4.03} & \textbf{4.37} & \textbf{4.97} & 3.33 & \textbf{5.00} & 4.90 \\
      \bottomrule
    \end{tabular}
    \vspace{-3mm}
  \end{table}

\vspace{-3mm}
\subsection{Out-of-domain Generalization Ability}
To answer \textbf{RQ2}, we compare model performance on unseen corpus under the zero-shot setting.
That is, given a model trained on XSUM, we evaluate its performance on the out-of-domain (OOD) test examples from CNN/DM and vice versa. 
Specifically, we sample $2000$ examples from each test set for evaluation.

As shown in Table \ref{tab:Comparisons_cross}, we observe that: though all the models struggle with OOD test examples, CI-Seq2Seq outperform baselines. For example, when training on XSUM and testing on CNN/DM, CI-Seq2Seq$^{pega}$ beats PEGASUS by 11.55\% in terms of RL.
These results demonstrate that capturing the invariant causal relationships can empower the summarization model with generalization ability.

\begin{table}[h]

  \renewcommand{\arraystretch}{1.1}
  \setlength\tabcolsep{2pt}
    \caption{OOD performance comparisons in terms of the generalization ability on unseen corpus. Best results are marked in boldface. * indicates statistically significant improvements over baselines (p-value < 0.05).}
    \label{tab:Comparisons_cross}
    \centering
    \begin{tabular}{ccccccc}
      \toprule
  \multirow{2}{*}{Train$\rightarrow $Test}  & \multicolumn{3}{c}{XSUM$\rightarrow $CNN/DM} & \multicolumn{3}{c}{CNN/DM$\rightarrow $XSUM} \\
      \cmidrule(r){2-4}
      \cmidrule(r){5-7}
      & R1 & R2 & RL & R1 & R2 & RL \\
      \midrule
      BART
      &25.10 &6.87 &17.86 &21.42 &3.78 &14.29 \\
      PEGASUS
      &28.14 &9.76 &19.83 &20.87 &3.88 &14.00 \\
      \hline
      CI-Seq2Seq$^{bart}$
      &25.49 &7.61 &18.32 &\textbf{23.92}* &\textbf{4.86}* &\textbf{15.52}* \\
      CI-Seq2Seq$^{pega}$
      &\textbf{30.37}* &\textbf{11.24}* &\textbf{22.12}* &21.75 &4.05 & 14.48\\
      \bottomrule
    \end{tabular}

  \end{table}

\begin{table*}[t]
  \small
  \renewcommand{\arraystretch}{0.9}
  \setlength\tabcolsep{5pt}
    \caption{An example (No.8) from the XSUM test data, to analyze the roles of content factors (CC and SC) and style factors (DS and SS). We mark the core content in blue and the side content in red.}
    \label{tab:case-main}
    \centering
    \scalebox{1}{
    \begin{tabularx}{\textwidth}{X}
      \toprule
      
      \textbf{Document:} ...The joint report from the Royal Society and Royal Academy of Engineering say the technique is safe if firms follow best practice and rules are enforced..."Our main conclusions are that the environmental risks of hydraulic fracturing for shale can be \textcolor{blue}{safely managed} provided there is best practice observed and provided it's enforced through \textcolor{blue}{strong regulation},"...\\
      \midrule
      \textbf{Ground-truth summary:} A gas extraction method which triggered two earth tremors near Blackpool last year \textcolor{blue}{should not cause earthquakes or contaminate water} but \textcolor{blue}{rules governing it will need tightening}, experts say.\\
      \midrule
      \textbf{BART:} Shale gas extraction can be carried out \textcolor{blue}{safely} in the UK, but \textcolor{blue}{stronger regulations} are needed to \textcolor{red}{protect public health}, a report says.\\
      \midrule
      \textbf{CI-Seq2Seq:} Shale gas extraction in the UK can be \textcolor{blue}{relatively safe}, but the government should \textcolor{blue}{strengthen regulations}, say scientists.\\
      \midrule
      \textbf{Attended$_{S}$:} (strengthening, environmental, shale), (strengthening, environmental, strong), (technique, safely, regulations), (technique, extraction, exploration), (technique, and, environmental), (being, moot, safely), (earth, small, tremors)...  \\ 
      \textbf{Attended$_{D}$:} (exploratory, fracking, involves), (involves, fracking, scientist), (fracking, acking, involves), (acking, atory, to), (involves, and, is), (involves, say, from), (and, or, into)...  \\
      \midrule
        \textbf{CR=0.1:} The environmental risks of fracking for shale gas in the UK are "very low", according to a new report.   \\
        \textbf{CR=0.4:} The use of fracking to extract shale gas in the UK can be safely done, according to a new report.  \\
        \textbf{CR=0.7:} Fracking, the controversial technique used to extract shale gas, can be safely done in the UK, according to a new report.  \\
      \bottomrule
    \end{tabularx}
    }
    \vspace{-2mm}
  \end{table*}

\vspace{-3mm}
\subsection{Interpretability of Latent Factors}
\label{sec:interpretable}

To answer \textbf{RQ3}, we analyze the roles of content factors and style factors through case study and visual analysis.

\begin{figure}[t]
  \centering
   \includegraphics[scale=0.235]{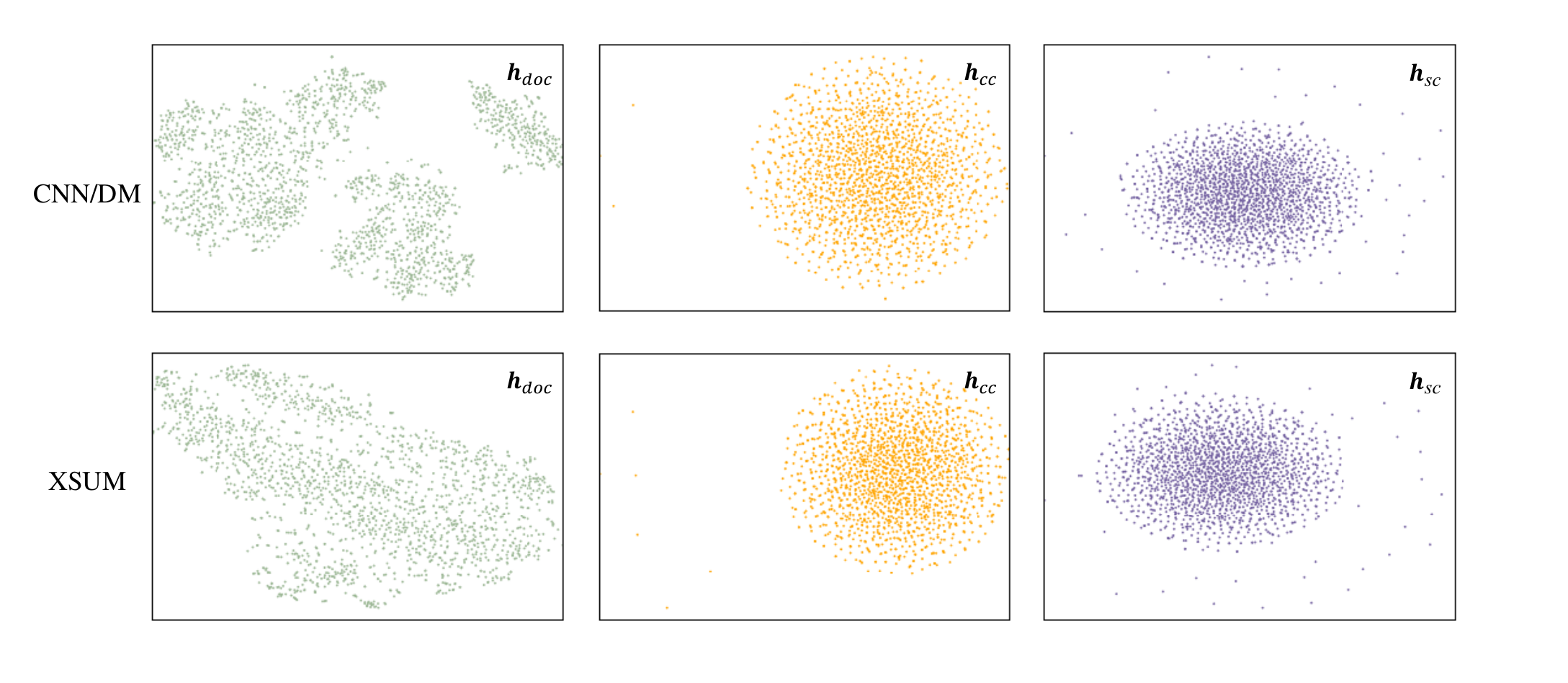}
   \caption{The t-SNE plot of $\textbf{h}_{doc}$,  $\textbf{h}_{cc}$ and $\textbf{h}_{sc}$ learned by ours. } 
   \label{fig:vis_emb}
\end{figure}

\noindent\textbf{Content Factors Analysis.}
To understand the influence of the CC and SC factors, we compare the top-3 attended words in the document when generating each token of the summary and document, based on the cross attention weights of Transformer.
As shown in Table \ref{tab:case-main}, summary generation guided by $\textbf{h}_{cc}$ prefers the tokens (\textbf{Attended}$_{S}$) conveying the core information of the document, e.g., ``shale'' and ``safely'', while document reconstruction guided by $\textbf{h}_{sc}$ and $\textbf{h}_{cc}$ attends to inessential words (\textbf{Attended}$_{D}$), e.g., ``involves'' and ``acking''. 
Without $\textbf{h}_{sc}$, the generated summary only captures the core content ``safe'' and ``strengthen regulations'', omitting the side content ``protect public health'' which exhibits frequent co-occurrence with ``safe'' in the corpus. 
This case indicates that the learned representations $\textbf{h}_{cc}$ and $\textbf{h}_{sc}$ can mimic the CC and SC factors to capture the core and side content in the document, respectively.
We also visually analyze the learned CC and SC representations. 
Specifically, we randomly sample 2000 test examples from XSUM and CNN/DM respectively, and then apply t-SNE \cite{van2008visualizing} to visualize $\textbf{h}_{doc}$, $\textbf{h}_{cc}$ and $\textbf{h}_{sc}$. 
As shown in Figure \ref{fig:vis_emb}, we can observe that: 
\begin{enumerate*}[label=(\roman*)]
    \item The distributions of both $\textbf{h}_{cc}$ and $\textbf{h}_{sc}$ are smoother than that of $\textbf{h}_{doc}$, indicating that by splitting the mixed information into distinct parts, each part will contain purer information.
    \item The distribution of $\textbf{h}_{cc}$ exhibits higher uniformity, whereas $\textbf{h}_{sc}$ demonstrates greater scattering. This observation indicates that the SC factors capture diverse side information for document generation and thus are dispersive.
\end{enumerate*}

\noindent\textbf{Style Factors Analysis.}
To understand the influence of the DS and SS factors, we vary $CR$ between them to explicitly control the summary generation.
Specifically, we vary $CR$ from 0.1 to 0.7 to change $\textbf{h}_{ss}$.
As shown in Table \ref{tab:case-main}, the generated summary is concise when $CR=0.1$, only including the necessary information, e.g., ``low'' and ``risks''.
When $CR = 0.7$, the generated summary contains more specific description, e.g., ``the controversial technique''.
The summary becomes more detailed as $CR$ increases.
Note that the goal of controlled generation here is not precise length control, but to control the style of the summary by utilizing $CR$ as weakly-supervised signal.
The results indicate that $\textbf{h}_{ss}$ can mimic the SS factors to actively control the writing style of the summary.

\vspace{-2mm}
\subsection{Impact of Latent Factors and Constraints}
\label{appendix-ablation}

To answer \textbf{RQ4}, we perform ablations on XSUM to analyze injection ways of latent factors as well as the necessity of confounder information and the content/style guidance served as constraints.

\textbf{Impact of Latent Factors.}
We removed the addition of $h_x/h_y$, and the replacement of the first token in the decoder, respectively.
As shown in the middle of Table \ref{tab:ablation}, we found that both addition and replacement operations contribute the prediction performance, and the replacement of the starting token matters more.

\textbf{Impact of Constraints.}
For confounder, we set $k_u=1$ to eliminate its information.
For content, we remove the content guidance loss in Equation \ref{eqn:lda-loss}.
For style, we sample $l_{ss}$ in the same way as $l_{ds}$ without additional bridge between them.
As shown in the bottom of Table \ref{tab:ablation}, removing constraints on either the confounder or content/style guidance hurts the prediction performance. This demonstrates the necessity of all constraints, which is consistent with our theory that they are essential for identifying latent factors.

\begin{table}[t]
  \renewcommand{\arraystretch}{0.85}
  \setlength\tabcolsep{8pt}
    \caption{Ablations of injection ways of latent factors \textbf{(middle)} as well as their constraints \textbf{(bottom)} on the subset of XSUM.}
    \label{tab:ablation}
    \centering
    \begin{tabular}{cccc}
      \toprule
      Model & R1 & R2 & RL  \\
      \midrule
      CI-Seq2Seq  &41.4  &17.75  &33.87  \\
      \midrule
      w/o addition  &40.97  &17.78  &33.77  \\
      w/o replacement  &36.00  &14.84  &28.74  \\
      \midrule
      w/o confounder information  &40.46  &17.60  &33.45  \\
      w/o content guidance  &40.87  &17.47  &33.14  \\
      w/o style guidance  &39.74  &16.55  &32.21  \\
      \bottomrule
    \end{tabular}
  \end{table}

\section{Conclusion}
In this paper, we presented a principled causal perspective for text summarization.
Theoretically, we proved the identifiability of the causal and non-causal factors in SCM to ensure these latent factors to be separated. 
Inspired by the identifiability theory, we proposed CI-Seq2Seq to learn causal representations that could mimic the causal factors for summary generation. 
We hope the paradigm can illuminate a promising technical direction of causality in NLP.  

One limitation of our method is the slightly higher computational cost than the original Seq2Seq architecture, due to the introduction of additional parameters and the optimization procedure during inference. 
To address this, we plan to reduce the dimension of latent representations and explore other optimization tools. 
We also want to explore diverse ways to utilize confounder information and define causal factors, which can better showcase our model's strengths under the identifiability guarantees. Besides, we are interested in inducing the causal structure into extractive summarization, and exploring the controllability on more aspects. 

\vspace{-2mm}
\begin{acks}
This work was funded by the National Natural Science Foundation of China (NSFC) under Grants No. 62006218, the China Scholarship Council under Grants No. 202104910234, the CAS Project for Young Scientists in Basic Research under Grant No. YSBR-034, the Innovation Project of ICT CAS under Grants No. E261090, the project under Grants No. JCKY2022130C039 and the Lenovo-CAS Joint Lab Youth Scientist Project.
\end{acks}

\balance
\bibliographystyle{ACM-Reference-Format}
\bibliography{reference}


\begin{thebibliography}{82}


\ifx \showCODEN    \undefined \def \showCODEN     #1{\unskip}     \fi
\ifx \showDOI      \undefined \def \showDOI       #1{#1}\fi
\ifx \showISBNx    \undefined \def \showISBNx     #1{\unskip}     \fi
\ifx \showISBNxiii \undefined \def \showISBNxiii  #1{\unskip}     \fi
\ifx \showISSN     \undefined \def \showISSN      #1{\unskip}     \fi
\ifx \showLCCN     \undefined \def \showLCCN      #1{\unskip}     \fi
\ifx \shownote     \undefined \def \shownote      #1{#1}          \fi
\ifx \showarticletitle \undefined \def \showarticletitle #1{#1}   \fi
\ifx \showURL      \undefined \def \showURL       {\relax}        \fi
\providecommand\bibfield[2]{#2}
\providecommand\bibinfo[2]{#2}
\providecommand\natexlab[1]{#1}
\providecommand\showeprint[2][]{arXiv:#2}

\bibitem[Ahuja et~al\mbox{.}(2022)]%
        {ahuja2022towards}
\bibfield{author}{\bibinfo{person}{Kartik Ahuja}, \bibinfo{person}{Divyat
  Mahajan}, \bibinfo{person}{Vasilis Syrgkanis}, {and} \bibinfo{person}{Ioannis
  Mitliagkas}.} \bibinfo{year}{2022}\natexlab{}.
\newblock \showarticletitle{Towards efficient representation identification in
  supervised learning}. In \bibinfo{booktitle}{\emph{CLeaR}}.
\newblock


\bibitem[Altman and Krzywinski(2015)]%
        {altman2015points}
\bibfield{author}{\bibinfo{person}{Naomi Altman} {and} \bibinfo{person}{Martin
  Krzywinski}.} \bibinfo{year}{2015}\natexlab{}.
\newblock \showarticletitle{Points of Significance: Association, correlation
  and causation.}
\newblock \bibinfo{journal}{\emph{Nature methods}} (\bibinfo{year}{2015}).
\newblock


\bibitem[Banko et~al\mbox{.}(2000)]%
        {banko2000headline}
\bibfield{author}{\bibinfo{person}{Michele Banko}, \bibinfo{person}{Vibhu~O.
  Mittal}, {and} \bibinfo{person}{Michael~J. Witbrock}.}
  \bibinfo{year}{2000}\natexlab{}.
\newblock \showarticletitle{Headline Generation Based on Statistical
  Translation}. In \bibinfo{booktitle}{\emph{ACL}}.
\newblock


\bibitem[Blei et~al\mbox{.}(2001)]%
        {blei2003latent}
\bibfield{author}{\bibinfo{person}{David~M. Blei}, \bibinfo{person}{Andrew~Y.
  Ng}, {and} \bibinfo{person}{Michael~I. Jordan}.}
  \bibinfo{year}{2001}\natexlab{}.
\newblock \showarticletitle{Latent Dirichlet Allocation}. In
  \bibinfo{booktitle}{\emph{NIPS}}.
\newblock


\bibitem[Bottou et~al\mbox{.}(2013)]%
        {bottou2013counterfactual}
\bibfield{author}{\bibinfo{person}{L{\'{e}}on Bottou}, \bibinfo{person}{Jonas
  Peters}, \bibinfo{person}{Joaquin~Qui{\~{n}}onero Candela},
  \bibinfo{person}{Denis~Xavier Charles}, \bibinfo{person}{Max Chickering},
  \bibinfo{person}{Elon Portugaly}, \bibinfo{person}{Dipankar Ray},
  \bibinfo{person}{Patrice~Y. Simard}, {and} \bibinfo{person}{Ed Snelson}.}
  \bibinfo{year}{2013}\natexlab{}.
\newblock \showarticletitle{Counterfactual reasoning and learning systems: the
  example of computational advertising}.
\newblock \bibinfo{journal}{\emph{J. Mach. Learn. Res.}} \bibinfo{volume}{14},
  \bibinfo{number}{1} (\bibinfo{year}{2013}).
\newblock


\bibitem[Cao et~al\mbox{.}(2022)]%
        {cao2022can}
\bibfield{author}{\bibinfo{person}{Boxi Cao}, \bibinfo{person}{Hongyu Lin},
  \bibinfo{person}{Xianpei Han}, \bibinfo{person}{Fangchao Liu}, {and}
  \bibinfo{person}{Le Sun}.} \bibinfo{year}{2022}\natexlab{}.
\newblock \showarticletitle{Can Prompt Probe Pretrained Language Models?
  Understanding the Invisible Risks from a Causal View}. In
  \bibinfo{booktitle}{\emph{ACL}}.
\newblock


\bibitem[Chen and Yang(2021)]%
        {chen2021simple}
\bibfield{author}{\bibinfo{person}{Jiaao Chen} {and} \bibinfo{person}{Diyi
  Yang}.} \bibinfo{year}{2021}\natexlab{}.
\newblock \showarticletitle{Simple Conversational Data Augmentation for
  Semi-supervised Abstractive Dialogue Summarization}. In
  \bibinfo{booktitle}{\emph{EMNLP}}.
\newblock


\bibitem[Chen et~al\mbox{.}(2021)]%
        {chen2021confounded}
\bibfield{author}{\bibinfo{person}{Wenqing Chen}, \bibinfo{person}{Jidong
  Tian}, \bibinfo{person}{Yitian Li}, \bibinfo{person}{Hao He}, {and}
  \bibinfo{person}{Yaohui Jin}.} \bibinfo{year}{2021}\natexlab{}.
\newblock \showarticletitle{De-Confounded Variational Encoder-Decoder for
  Logical Table-to-Text Generation}. In \bibinfo{booktitle}{\emph{ACL-IJCNLP}}.
\newblock


\bibitem[Chen et~al\mbox{.}(2020)]%
        {chen2020exploring}
\bibfield{author}{\bibinfo{person}{Wenqing Chen}, \bibinfo{person}{Jidong
  Tian}, \bibinfo{person}{Liqiang Xiao}, \bibinfo{person}{Hao He}, {and}
  \bibinfo{person}{Yaohui Jin}.} \bibinfo{year}{2020}\natexlab{}.
\newblock \showarticletitle{Exploring Logically Dependent Multi-task Learning
  with Causal Inference}. In \bibinfo{booktitle}{\emph{EMNLP}}.
\newblock


\bibitem[Choi et~al\mbox{.}(2019)]%
        {choi-etal-2019-vae}
\bibfield{author}{\bibinfo{person}{Hyungtak Choi}, \bibinfo{person}{Lohith
  Ravuru}, \bibinfo{person}{Tomasz Dryja{\'n}ski}, \bibinfo{person}{Sunghan
  Rye}, \bibinfo{person}{Donghyun Lee}, \bibinfo{person}{Hojung Lee}, {and}
  \bibinfo{person}{Inchul Hwang}.} \bibinfo{year}{2019}\natexlab{}.
\newblock \showarticletitle{{VAE}-{PGN} based Abstractive Model in Multi-stage
  Architecture for Text Summarization}. In \bibinfo{booktitle}{\emph{INLG}}.
\newblock


\bibitem[Dougrez-Lewis et~al\mbox{.}(2021)]%
        {dougrez2021learning}
\bibfield{author}{\bibinfo{person}{John Dougrez-Lewis}, \bibinfo{person}{Maria
  Liakata}, \bibinfo{person}{Elena Kochkina}, {and} \bibinfo{person}{Yulan
  He}.} \bibinfo{year}{2021}\natexlab{}.
\newblock \showarticletitle{Learning Disentangled Latent Topics for {T}witter
  Rumour Veracity Classification}. In \bibinfo{booktitle}{\emph{ACL-IJCNLP}}.
\newblock


\bibitem[Du et~al\mbox{.}(2022)]%
        {du-etal-2022-glm}
\bibfield{author}{\bibinfo{person}{Zhengxiao Du}, \bibinfo{person}{Yujie Qian},
  \bibinfo{person}{Xiao Liu}, \bibinfo{person}{Ming Ding},
  \bibinfo{person}{Jiezhong Qiu}, \bibinfo{person}{Zhilin Yang}, {and}
  \bibinfo{person}{Jie Tang}.} \bibinfo{year}{2022}\natexlab{}.
\newblock \showarticletitle{{GLM}: General Language Model Pretraining with
  Autoregressive Blank Infilling}. In \bibinfo{booktitle}{\emph{ACL}}.
\newblock


\bibitem[Fabbri et~al\mbox{.}(2021)]%
        {fabbri2021summeval}
\bibfield{author}{\bibinfo{person}{Alexander~R. Fabbri},
  \bibinfo{person}{Wojciech Kry{\'s}ci{\'n}ski}, \bibinfo{person}{Bryan
  McCann}, \bibinfo{person}{Caiming Xiong}, \bibinfo{person}{Richard Socher},
  {and} \bibinfo{person}{Dragomir Radev}.} \bibinfo{year}{2021}\natexlab{}.
\newblock \showarticletitle{{S}umm{E}val: Re-evaluating Summarization
  Evaluation}.
\newblock \bibinfo{journal}{\emph{Transactions of the Association for
  Computational Linguistics}}  \bibinfo{volume}{9} (\bibinfo{year}{2021}).
\newblock


\bibitem[Fan et~al\mbox{.}(2018)]%
        {fan2018controllable}
\bibfield{author}{\bibinfo{person}{Angela Fan}, \bibinfo{person}{David
  Grangier}, {and} \bibinfo{person}{Michael Auli}.}
  \bibinfo{year}{2018}\natexlab{}.
\newblock \showarticletitle{Controllable Abstractive Summarization}. In
  \bibinfo{booktitle}{\emph{Proceedings of the 2nd Workshop on Neural Machine
  Translation and Generation}}.
\newblock


\bibitem[Fu et~al\mbox{.}(2020)]%
        {fu2020document}
\bibfield{author}{\bibinfo{person}{Xiyan Fu}, \bibinfo{person}{Jun Wang},
  \bibinfo{person}{Jinghan Zhang}, \bibinfo{person}{Jinmao Wei}, {and}
  \bibinfo{person}{Zhenglu Yang}.} \bibinfo{year}{2020}\natexlab{}.
\newblock \showarticletitle{Document Summarization with {VHTM:} Variational
  Hierarchical Topic-Aware Mechanism}. In
  \bibinfo{booktitle}{\emph{AAAI-EAAI}}.
\newblock


\bibitem[Gehrmann et~al\mbox{.}(2018)]%
        {gehrmann2018bottom}
\bibfield{author}{\bibinfo{person}{Sebastian Gehrmann},
  \bibinfo{person}{Yuntian Deng}, {and} \bibinfo{person}{Alexander Rush}.}
  \bibinfo{year}{2018}\natexlab{}.
\newblock \showarticletitle{Bottom-Up Abstractive Summarization}. In
  \bibinfo{booktitle}{\emph{EMNLP}}.
\newblock


\bibitem[Glymour et~al\mbox{.}(2016)]%
        {glymour2016causal}
\bibfield{author}{\bibinfo{person}{Madelyn Glymour}, \bibinfo{person}{Judea
  Pearl}, {and} \bibinfo{person}{Nicholas~P Jewell}.}
  \bibinfo{year}{2016}\natexlab{}.
\newblock \bibinfo{booktitle}{\emph{Causal inference in statistics: A primer}}.
\newblock \bibinfo{publisher}{John Wiley \& Sons}.
\newblock


\bibitem[Hahn and Mani(2000)]%
        {hahn2000challenges}
\bibfield{author}{\bibinfo{person}{Udo Hahn} {and} \bibinfo{person}{Inderjeet
  Mani}.} \bibinfo{year}{2000}\natexlab{}.
\newblock \showarticletitle{The Challenges of Automatic Summarization}.
\newblock \bibinfo{journal}{\emph{Computer}} \bibinfo{volume}{33},
  \bibinfo{number}{11} (\bibinfo{year}{2000}).
\newblock


\bibitem[Hermann et~al\mbox{.}(2015)]%
        {hermann2015teaching}
\bibfield{author}{\bibinfo{person}{Karl~Moritz Hermann},
  \bibinfo{person}{Tom{\'{a}}s Kocisk{\'{y}}}, \bibinfo{person}{Edward
  Grefenstette}, \bibinfo{person}{Lasse Espeholt}, \bibinfo{person}{Will Kay},
  \bibinfo{person}{Mustafa Suleyman}, {and} \bibinfo{person}{Phil Blunsom}.}
  \bibinfo{year}{2015}\natexlab{}.
\newblock \showarticletitle{Teaching Machines to Read and Comprehend}. In
  \bibinfo{booktitle}{\emph{NIPS}}.
\newblock


\bibitem[Hsu et~al\mbox{.}(2017)]%
        {hsu2017unsupervised}
\bibfield{author}{\bibinfo{person}{Wei{-}Ning Hsu}, \bibinfo{person}{Yu Zhang},
  {and} \bibinfo{person}{James~R. Glass}.} \bibinfo{year}{2017}\natexlab{}.
\newblock \showarticletitle{Unsupervised Learning of Disentangled and
  Interpretable Representations from Sequential Data}. In
  \bibinfo{booktitle}{\emph{NeurIPS}}.
\newblock


\bibitem[Hu and Li(2021)]%
        {hu2021causal}
\bibfield{author}{\bibinfo{person}{Zhiting Hu} {and} \bibinfo{person}{Li~Erran
  Li}.} \bibinfo{year}{2021}\natexlab{}.
\newblock \showarticletitle{A Causal Lens for Controllable Text Generation}. In
  \bibinfo{booktitle}{\emph{NeurIPS}}.
\newblock


\bibitem[Huang et~al\mbox{.}(2020)]%
        {huang2020knowledge}
\bibfield{author}{\bibinfo{person}{Luyang Huang}, \bibinfo{person}{Lingfei Wu},
  {and} \bibinfo{person}{Lu Wang}.} \bibinfo{year}{2020}\natexlab{}.
\newblock \showarticletitle{Knowledge Graph-Augmented Abstractive Summarization
  with Semantic-Driven Cloze Reward}. In \bibinfo{booktitle}{\emph{ACL}}.
\newblock


\bibitem[Janzing et~al\mbox{.}(2009)]%
        {JanzingPMS09}
\bibfield{author}{\bibinfo{person}{Dominik Janzing}, \bibinfo{person}{Jonas
  Peters}, \bibinfo{person}{Joris~M. Mooij}, {and} \bibinfo{person}{Bernhard
  Sch{\"{o}}lkopf}.} \bibinfo{year}{2009}\natexlab{}.
\newblock \showarticletitle{Identifying confounders using additive noise
  models}. In \bibinfo{booktitle}{\emph{UAI}}.
\newblock


\bibitem[Joachims et~al\mbox{.}(2017)]%
        {joachims2017unbiased}
\bibfield{author}{\bibinfo{person}{Thorsten Joachims}, \bibinfo{person}{Adith
  Swaminathan}, {and} \bibinfo{person}{Tobias Schnabel}.}
  \bibinfo{year}{2017}\natexlab{}.
\newblock \showarticletitle{Unbiased Learning-to-Rank with Biased Feedback}. In
  \bibinfo{booktitle}{\emph{WSDM}}.
\newblock


\bibitem[Khemakhem et~al\mbox{.}(2020)]%
        {khemakhem2020variational}
\bibfield{author}{\bibinfo{person}{Ilyes Khemakhem},
  \bibinfo{person}{Diederik~P. Kingma}, \bibinfo{person}{Ricardo~Pio Monti},
  {and} \bibinfo{person}{Aapo Hyv{\"{a}}rinen}.}
  \bibinfo{year}{2020}\natexlab{}.
\newblock \showarticletitle{Variational Autoencoders and Nonlinear {ICA:} {A}
  Unifying Framework}. In \bibinfo{booktitle}{\emph{AISTATS}},
  Vol.~\bibinfo{volume}{108}.
\newblock


\bibitem[Kingma and Welling(2014)]%
        {kingma2013auto}
\bibfield{author}{\bibinfo{person}{Diederik~P. Kingma} {and}
  \bibinfo{person}{Max Welling}.} \bibinfo{year}{2014}\natexlab{}.
\newblock \showarticletitle{Auto-Encoding Variational Bayes}. In
  \bibinfo{booktitle}{\emph{ICLR}}.
\newblock


\bibitem[Kryscinski et~al\mbox{.}(2019)]%
        {kryscinski2019neural}
\bibfield{author}{\bibinfo{person}{Wojciech Kryscinski},
  \bibinfo{person}{Nitish~Shirish Keskar}, \bibinfo{person}{Bryan McCann},
  \bibinfo{person}{Caiming Xiong}, {and} \bibinfo{person}{Richard Socher}.}
  \bibinfo{year}{2019}\natexlab{}.
\newblock \showarticletitle{Neural Text Summarization: A Critical Evaluation}.
  In \bibinfo{booktitle}{\emph{EMNLP-IJCNLP}}.
\newblock


\bibitem[Lewis et~al\mbox{.}(2020)]%
        {lewis-etal-2020-bart}
\bibfield{author}{\bibinfo{person}{Mike Lewis}, \bibinfo{person}{Yinhan Liu},
  \bibinfo{person}{Naman Goyal}, \bibinfo{person}{Marjan Ghazvininejad},
  \bibinfo{person}{Abdelrahman Mohamed}, \bibinfo{person}{Omer Levy},
  \bibinfo{person}{Veselin Stoyanov}, {and} \bibinfo{person}{Luke
  Zettlemoyer}.} \bibinfo{year}{2020}\natexlab{}.
\newblock \showarticletitle{{BART}: Denoising Sequence-to-Sequence Pre-training
  for Natural Language Generation, Translation, and Comprehension}. In
  \bibinfo{booktitle}{\emph{ACL}}.
\newblock


\bibitem[Li et~al\mbox{.}(2020)]%
        {li2020optimus}
\bibfield{author}{\bibinfo{person}{Chunyuan Li}, \bibinfo{person}{Xiang Gao},
  \bibinfo{person}{Yuan Li}, \bibinfo{person}{Baolin Peng},
  \bibinfo{person}{Xiujun Li}, \bibinfo{person}{Yizhe Zhang}, {and}
  \bibinfo{person}{Jianfeng Gao}.} \bibinfo{year}{2020}\natexlab{}.
\newblock \showarticletitle{Optimus: Organizing Sentences via Pre-trained
  Modeling of a Latent Space}. In \bibinfo{booktitle}{\emph{EMNLP}}.
\newblock


\bibitem[Lin(2004)]%
        {lin2004rouge}
\bibfield{author}{\bibinfo{person}{Chin-Yew Lin}.}
  \bibinfo{year}{2004}\natexlab{}.
\newblock \showarticletitle{{ROUGE}: A Package for Automatic Evaluation of
  Summaries}. In \bibinfo{booktitle}{\emph{Text Summarization Branches Out}}.
\newblock


\bibitem[Liu et~al\mbox{.}(2021)]%
        {liu2021learning}
\bibfield{author}{\bibinfo{person}{Chang Liu}, \bibinfo{person}{Xinwei Sun},
  \bibinfo{person}{Jindong Wang}, \bibinfo{person}{Haoyue Tang},
  \bibinfo{person}{Tao Li}, \bibinfo{person}{Tao Qin}, \bibinfo{person}{Wei
  Chen}, {and} \bibinfo{person}{Tie{-}Yan Liu}.}
  \bibinfo{year}{2021}\natexlab{}.
\newblock \showarticletitle{Learning Causal Semantic Representation for
  Out-of-Distribution Prediction}. In \bibinfo{booktitle}{\emph{NeurIPS}}.
\newblock


\bibitem[Liu et~al\mbox{.}(2015)]%
        {liu2015toward}
\bibfield{author}{\bibinfo{person}{Fei Liu}, \bibinfo{person}{Jeffrey
  Flanigan}, \bibinfo{person}{Sam Thomson}, \bibinfo{person}{Norman Sadeh},
  {and} \bibinfo{person}{Noah~A. Smith}.} \bibinfo{year}{2015}\natexlab{}.
\newblock \showarticletitle{Toward Abstractive Summarization Using Semantic
  Representations}. In \bibinfo{booktitle}{\emph{NAACL}}.
\newblock


\bibitem[Liu et~al\mbox{.}(2018)]%
        {liu2018generative}
\bibfield{author}{\bibinfo{person}{Linqing Liu}, \bibinfo{person}{Yao Lu},
  \bibinfo{person}{Min Yang}, \bibinfo{person}{Qiang Qu}, \bibinfo{person}{Jia
  Zhu}, {and} \bibinfo{person}{Hongyan Li}.} \bibinfo{year}{2018}\natexlab{}.
\newblock \showarticletitle{Generative Adversarial Network for Abstractive Text
  Summarization}. In \bibinfo{booktitle}{\emph{AAAI-EAAI}}.
\newblock


\bibitem[Liu et~al\mbox{.}(2022)]%
        {liu-etal-2022-length}
\bibfield{author}{\bibinfo{person}{Yizhu Liu}, \bibinfo{person}{Qi Jia}, {and}
  \bibinfo{person}{Kenny Zhu}.} \bibinfo{year}{2022}\natexlab{}.
\newblock \showarticletitle{Length Control in Abstractive Summarization by
  Pretraining Information Selection}. In \bibinfo{booktitle}{\emph{ACL}}.
\newblock


\bibitem[Lu et~al\mbox{.}(2022)]%
        {lu2021invariant}
\bibfield{author}{\bibinfo{person}{Chaochao Lu}, \bibinfo{person}{Yuhuai Wu},
  \bibinfo{person}{Jos{\'{e}}~Miguel Hern{\'{a}}ndez{-}Lobato}, {and}
  \bibinfo{person}{Bernhard Sch{\"{o}}lkopf}.} \bibinfo{year}{2022}\natexlab{}.
\newblock \showarticletitle{Invariant Causal Representation Learning for
  Out-of-Distribution Generalization}. In \bibinfo{booktitle}{\emph{ICLR}}.
\newblock


\bibitem[Maybury(1999)]%
        {maybury1999advances}
\bibfield{author}{\bibinfo{person}{Mani Maybury}.}
  \bibinfo{year}{1999}\natexlab{}.
\newblock \bibinfo{booktitle}{\emph{Advances in automatic text summarization}}.
\newblock \bibinfo{publisher}{MIT press}.
\newblock


\bibitem[Mikolov et~al\mbox{.}(2013)]%
        {mikolov2013distributed}
\bibfield{author}{\bibinfo{person}{Tom{\'{a}}s Mikolov}, \bibinfo{person}{Ilya
  Sutskever}, \bibinfo{person}{Kai Chen}, \bibinfo{person}{Gregory~S. Corrado},
  {and} \bibinfo{person}{Jeffrey Dean}.} \bibinfo{year}{2013}\natexlab{}.
\newblock \showarticletitle{Distributed Representations of Words and Phrases
  and their Compositionality}. In \bibinfo{booktitle}{\emph{NIPS}}.
\newblock


\bibitem[Mitrovic et~al\mbox{.}(2021)]%
        {mitrovicrepresentation}
\bibfield{author}{\bibinfo{person}{Jovana Mitrovic}, \bibinfo{person}{Brian
  McWilliams}, \bibinfo{person}{Jacob~C. Walker}, \bibinfo{person}{Lars~Holger
  Buesing}, {and} \bibinfo{person}{Charles Blundell}.}
  \bibinfo{year}{2021}\natexlab{}.
\newblock \showarticletitle{Representation Learning via Invariant Causal
  Mechanisms}. In \bibinfo{booktitle}{\emph{ICLR}}.
\newblock


\bibitem[Moraffah et~al\mbox{.}(2020)]%
        {moraffah2020causal}
\bibfield{author}{\bibinfo{person}{Raha Moraffah}, \bibinfo{person}{Mansooreh
  Karami}, \bibinfo{person}{Ruocheng Guo}, \bibinfo{person}{Adrienne Raglin},
  {and} \bibinfo{person}{Huan Liu}.} \bibinfo{year}{2020}\natexlab{}.
\newblock \showarticletitle{Causal Interpretability for Machine Learning -
  Problems, Methods and Evaluation}.
\newblock \bibinfo{journal}{\emph{{SIGKDD} Explor.}} \bibinfo{volume}{22},
  \bibinfo{number}{1} (\bibinfo{year}{2020}).
\newblock


\bibitem[Nallapati et~al\mbox{.}(2017)]%
        {nallapati2017summarunner}
\bibfield{author}{\bibinfo{person}{Ramesh Nallapati}, \bibinfo{person}{Feifei
  Zhai}, {and} \bibinfo{person}{Bowen Zhou}.} \bibinfo{year}{2017}\natexlab{}.
\newblock \showarticletitle{SummaRuNNer: {A} Recurrent Neural Network Based
  Sequence Model for Extractive Summarization of Documents}. In
  \bibinfo{booktitle}{\emph{AAAI}}.
\newblock


\bibitem[Nan et~al\mbox{.}(2021)]%
        {nan2021uncovering}
\bibfield{author}{\bibinfo{person}{Guoshun Nan}, \bibinfo{person}{Jiaqi Zeng},
  \bibinfo{person}{Rui Qiao}, \bibinfo{person}{Zhijiang Guo}, {and}
  \bibinfo{person}{Wei Lu}.} \bibinfo{year}{2021}\natexlab{}.
\newblock \showarticletitle{Uncovering Main Causalities for Long-tailed
  Information Extraction}. In \bibinfo{booktitle}{\emph{EMNLP}}.
\newblock


\bibitem[Nangi et~al\mbox{.}(2021)]%
        {nangi2021counterfactuals}
\bibfield{author}{\bibinfo{person}{Sharmila~Reddy Nangi},
  \bibinfo{person}{Niyati Chhaya}, \bibinfo{person}{Sopan Khosla},
  \bibinfo{person}{Nikhil Kaushik}, {and} \bibinfo{person}{Harshit Nyati}.}
  \bibinfo{year}{2021}\natexlab{}.
\newblock \showarticletitle{Counterfactuals to Control Latent Disentangled Text
  Representations for Style Transfer}. In
  \bibinfo{booktitle}{\emph{ACL-IJCNLP}}.
\newblock


\bibitem[Narayan et~al\mbox{.}(2018a)]%
        {narayan2018don}
\bibfield{author}{\bibinfo{person}{Shashi Narayan}, \bibinfo{person}{Shay~B.
  Cohen}, {and} \bibinfo{person}{Mirella Lapata}.}
  \bibinfo{year}{2018}\natexlab{a}.
\newblock \showarticletitle{Don{'}t Give Me the Details, Just the Summary!
  Topic-Aware Convolutional Neural Networks for Extreme Summarization}. In
  \bibinfo{booktitle}{\emph{EMNLP}}.
\newblock


\bibitem[Narayan et~al\mbox{.}(2018b)]%
        {narayan2018ranking}
\bibfield{author}{\bibinfo{person}{Shashi Narayan}, \bibinfo{person}{Shay~B.
  Cohen}, {and} \bibinfo{person}{Mirella Lapata}.}
  \bibinfo{year}{2018}\natexlab{b}.
\newblock \showarticletitle{Ranking Sentences for Extractive Summarization with
  Reinforcement Learning}. In \bibinfo{booktitle}{\emph{NAACL}}.
\newblock


\bibitem[Nenkova and McKeown(2012)]%
        {nenkova2012survey}
\bibfield{author}{\bibinfo{person}{Ani Nenkova} {and} \bibinfo{person}{Kathleen
  McKeown}.} \bibinfo{year}{2012}\natexlab{}.
\newblock \showarticletitle{A survey of text summarization techniques}.
\newblock In \bibinfo{booktitle}{\emph{Mining text data}}.
  \bibinfo{publisher}{Springer}.
\newblock


\bibitem[Niu et~al\mbox{.}(2021)]%
        {niu2021counterfactual}
\bibfield{author}{\bibinfo{person}{Yulei Niu}, \bibinfo{person}{Kaihua Tang},
  \bibinfo{person}{Hanwang Zhang}, \bibinfo{person}{Zhiwu Lu},
  \bibinfo{person}{Xian{-}Sheng Hua}, {and} \bibinfo{person}{Ji{-}Rong Wen}.}
  \bibinfo{year}{2021}\natexlab{}.
\newblock \showarticletitle{Counterfactual {VQA:} {A} Cause-Effect Look at
  Language Bias}. In \bibinfo{booktitle}{\emph{CVPR}}.
\newblock


\bibitem[Paulus et~al\mbox{.}(2018)]%
        {paulus2018deep}
\bibfield{author}{\bibinfo{person}{Romain Paulus}, \bibinfo{person}{Caiming
  Xiong}, {and} \bibinfo{person}{Richard Socher}.}
  \bibinfo{year}{2018}\natexlab{}.
\newblock \showarticletitle{A Deep Reinforced Model for Abstractive
  Summarization}. In \bibinfo{booktitle}{\emph{ICLR}}.
\newblock


\bibitem[Pearl(1995)]%
        {pearl1995causal}
\bibfield{author}{\bibinfo{person}{Judea Pearl}.}
  \bibinfo{year}{1995}\natexlab{}.
\newblock \showarticletitle{Causal diagrams for empirical research}.
\newblock \bibinfo{journal}{\emph{Biometrika}} (\bibinfo{year}{1995}).
\newblock


\bibitem[Pearl(2009a)]%
        {pearl2009causal}
\bibfield{author}{\bibinfo{person}{Judea Pearl}.}
  \bibinfo{year}{2009}\natexlab{a}.
\newblock \showarticletitle{Causal inference in statistics: An overview}.
\newblock \bibinfo{journal}{\emph{Statistics surveys}} (\bibinfo{year}{2009}).
\newblock


\bibitem[Pearl(2009b)]%
        {pearl2009causality}
\bibfield{author}{\bibinfo{person}{Judea Pearl}.}
  \bibinfo{year}{2009}\natexlab{b}.
\newblock \bibinfo{booktitle}{\emph{Causality}}.
\newblock \bibinfo{publisher}{Cambridge university press}.
\newblock


\bibitem[Pearl et~al\mbox{.}(2000)]%
        {pearl2000models}
\bibfield{author}{\bibinfo{person}{Judea Pearl} {et~al\mbox{.}}}
  \bibinfo{year}{2000}\natexlab{}.
\newblock \showarticletitle{Models, reasoning and inference}.
\newblock \bibinfo{journal}{\emph{Cambridge, UK: CambridgeUniversityPress}}
  (\bibinfo{year}{2000}).
\newblock


\bibitem[Peters et~al\mbox{.}(2017)]%
        {peters2017elements}
\bibfield{author}{\bibinfo{person}{Jonas Peters}, \bibinfo{person}{Dominik
  Janzing}, {and} \bibinfo{person}{Bernhard Sch{\"o}lkopf}.}
  \bibinfo{year}{2017}\natexlab{}.
\newblock \bibinfo{booktitle}{\emph{Elements of causal inference: foundations
  and learning algorithms}}.
\newblock \bibinfo{publisher}{The MIT Press}.
\newblock


\bibitem[Qian et~al\mbox{.}(2021)]%
        {qian2021counterfactual}
\bibfield{author}{\bibinfo{person}{Chen Qian}, \bibinfo{person}{Fuli Feng},
  \bibinfo{person}{Lijie Wen}, \bibinfo{person}{Chunping Ma}, {and}
  \bibinfo{person}{Pengjun Xie}.} \bibinfo{year}{2021}\natexlab{}.
\newblock \showarticletitle{Counterfactual Inference for Text Classification
  Debiasing}. In \bibinfo{booktitle}{\emph{ACL-IJCNLP}}.
\newblock


\bibitem[Raffel et~al\mbox{.}(2020)]%
        {raffel2020exploring}
\bibfield{author}{\bibinfo{person}{Colin Raffel}, \bibinfo{person}{Noam
  Shazeer}, \bibinfo{person}{Adam Roberts}, \bibinfo{person}{Katherine Lee},
  \bibinfo{person}{Sharan Narang}, \bibinfo{person}{Michael Matena},
  \bibinfo{person}{Yanqi Zhou}, \bibinfo{person}{Wei Li}, {and}
  \bibinfo{person}{Peter~J. Liu}.} \bibinfo{year}{2020}\natexlab{}.
\newblock \showarticletitle{Exploring the Limits of Transfer Learning with a
  Unified Text-to-Text Transformer}.
\newblock \bibinfo{journal}{\emph{J. Mach. Learn. Res.}}  \bibinfo{volume}{21}
  (\bibinfo{year}{2020}).
\newblock


\bibitem[Ruan et~al\mbox{.}(2022)]%
        {ruan2022histruct+}
\bibfield{author}{\bibinfo{person}{Qian Ruan}, \bibinfo{person}{Malte
  Ostendorff}, {and} \bibinfo{person}{Georg Rehm}.}
  \bibinfo{year}{2022}\natexlab{}.
\newblock \showarticletitle{{H}i{S}truct+: Improving Extractive Text
  Summarization with Hierarchical Structure Information}. In
  \bibinfo{booktitle}{\emph{ACL}}.
\newblock


\bibitem[Rubin(1974)]%
        {rubin1974estimating}
\bibfield{author}{\bibinfo{person}{Donald~B Rubin}.}
  \bibinfo{year}{1974}\natexlab{}.
\newblock \showarticletitle{Estimating causal effects of treatments in
  randomized and nonrandomized studies.}
\newblock \bibinfo{journal}{\emph{Journal of educational Psychology}}
  (\bibinfo{year}{1974}).
\newblock


\bibitem[Rush et~al\mbox{.}(2015)]%
        {rush2015neural}
\bibfield{author}{\bibinfo{person}{Alexander~M. Rush}, \bibinfo{person}{Sumit
  Chopra}, {and} \bibinfo{person}{Jason Weston}.}
  \bibinfo{year}{2015}\natexlab{}.
\newblock \showarticletitle{A Neural Attention Model for Abstractive Sentence
  Summarization}. In \bibinfo{booktitle}{\emph{EMNLP}}.
\newblock


\bibitem[Sch{\"o}lkopf(2022)]%
        {scholkopf2022causality}
\bibfield{author}{\bibinfo{person}{Bernhard Sch{\"o}lkopf}.}
  \bibinfo{year}{2022}\natexlab{}.
\newblock \showarticletitle{Causality for machine learning}.
\newblock In \bibinfo{booktitle}{\emph{Probabilistic and Causal Inference: The
  Works of Judea Pearl}}.
\newblock


\bibitem[Sch{\"{o}}lkopf et~al\mbox{.}(2012)]%
        {ScholkopfJPSZM12}
\bibfield{author}{\bibinfo{person}{Bernhard Sch{\"{o}}lkopf},
  \bibinfo{person}{Dominik Janzing}, \bibinfo{person}{Jonas Peters},
  \bibinfo{person}{Eleni Sgouritsa}, \bibinfo{person}{Kun Zhang}, {and}
  \bibinfo{person}{Joris~M. Mooij}.} \bibinfo{year}{2012}\natexlab{}.
\newblock \showarticletitle{On causal and anticausal learning}. In
  \bibinfo{booktitle}{\emph{ICML}}.
\newblock


\bibitem[Sun et~al\mbox{.}(2021)]%
        {sun2021recovering}
\bibfield{author}{\bibinfo{person}{Xinwei Sun}, \bibinfo{person}{Botong Wu},
  \bibinfo{person}{Xiangyu Zheng}, \bibinfo{person}{Chang Liu},
  \bibinfo{person}{Wei Chen}, \bibinfo{person}{Tao Qin}, {and}
  \bibinfo{person}{Tie{-}Yan Liu}.} \bibinfo{year}{2021}\natexlab{}.
\newblock \showarticletitle{Recovering Latent Causal Factor for Generalization
  to Distributional Shifts}. In \bibinfo{booktitle}{\emph{NeurIPS}}.
\newblock


\bibitem[Van~der Maaten and Hinton(2008)]%
        {van2008visualizing}
\bibfield{author}{\bibinfo{person}{Laurens Van~der Maaten} {and}
  \bibinfo{person}{Geoffrey Hinton}.} \bibinfo{year}{2008}\natexlab{}.
\newblock \showarticletitle{Visualizing data using t-SNE.}
\newblock \bibinfo{journal}{\emph{JMLR}} (\bibinfo{year}{2008}).
\newblock


\bibitem[Veitch et~al\mbox{.}(2021)]%
        {veitch2021counterfactual}
\bibfield{author}{\bibinfo{person}{Victor Veitch}, \bibinfo{person}{Alexander
  D'Amour}, \bibinfo{person}{Steve Yadlowsky}, {and} \bibinfo{person}{Jacob
  Eisenstein}.} \bibinfo{year}{2021}\natexlab{}.
\newblock \showarticletitle{Counterfactual Invariance to Spurious Correlations
  in Text Classification}. In \bibinfo{booktitle}{\emph{NeurIPS}}.
\newblock


\bibitem[Wang et~al\mbox{.}(2019)]%
        {wang2019self}
\bibfield{author}{\bibinfo{person}{Hong Wang}, \bibinfo{person}{Xin Wang},
  \bibinfo{person}{Wenhan Xiong}, \bibinfo{person}{Mo Yu},
  \bibinfo{person}{Xiaoxiao Guo}, \bibinfo{person}{Shiyu Chang}, {and}
  \bibinfo{person}{William~Yang Wang}.} \bibinfo{year}{2019}\natexlab{}.
\newblock \showarticletitle{Self-Supervised Learning for Contextualized
  Extractive Summarization}. In \bibinfo{booktitle}{\emph{ACL}}.
\newblock


\bibitem[Wang et~al\mbox{.}(2022)]%
        {wang2022causal}
\bibfield{author}{\bibinfo{person}{Wenjie Wang}, \bibinfo{person}{Xinyu Lin},
  \bibinfo{person}{Fuli Feng}, \bibinfo{person}{Xiangnan He},
  \bibinfo{person}{Min Lin}, {and} \bibinfo{person}{Tat{-}Seng Chua}.}
  \bibinfo{year}{2022}\natexlab{}.
\newblock \showarticletitle{Causal Representation Learning for
  Out-of-Distribution Recommendation}. In \bibinfo{booktitle}{\emph{WWW}}.
\newblock


\bibitem[Wei et~al\mbox{.}(2021)]%
        {wei2021model}
\bibfield{author}{\bibinfo{person}{Tianxin Wei}, \bibinfo{person}{Fuli Feng},
  \bibinfo{person}{Jiawei Chen}, \bibinfo{person}{Ziwei Wu},
  \bibinfo{person}{Jinfeng Yi}, {and} \bibinfo{person}{Xiangnan He}.}
  \bibinfo{year}{2021}\natexlab{}.
\newblock \showarticletitle{Model-Agnostic Counterfactual Reasoning for
  Eliminating Popularity Bias in Recommender System}. In
  \bibinfo{booktitle}{\emph{KDD}}.
\newblock


\bibitem[Xia et~al\mbox{.}(2020)]%
        {xia2020composed}
\bibfield{author}{\bibinfo{person}{Congying Xia}, \bibinfo{person}{Caiming
  Xiong}, \bibinfo{person}{Philip Yu}, {and} \bibinfo{person}{Richard Socher}.}
  \bibinfo{year}{2020}\natexlab{}.
\newblock \showarticletitle{Composed Variational Natural Language Generation
  for Few-shot Intents}. In \bibinfo{booktitle}{\emph{EMNLP}}.
\newblock


\bibitem[Xiao and Wang(2021)]%
        {xiao2021hallucination}
\bibfield{author}{\bibinfo{person}{Yijun Xiao} {and}
  \bibinfo{person}{William~Yang Wang}.} \bibinfo{year}{2021}\natexlab{}.
\newblock \showarticletitle{On Hallucination and Predictive Uncertainty in
  Conditional Language Generation}. In \bibinfo{booktitle}{\emph{EACL}}.
\newblock


\bibitem[Xie et~al\mbox{.}(2021)]%
        {xie2021factual}
\bibfield{author}{\bibinfo{person}{Yuexiang Xie}, \bibinfo{person}{Fei Sun},
  \bibinfo{person}{Yang Deng}, \bibinfo{person}{Yaliang Li}, {and}
  \bibinfo{person}{Bolin Ding}.} \bibinfo{year}{2021}\natexlab{}.
\newblock \showarticletitle{Factual Consistency Evaluation for Text
  Summarization via Counterfactual Estimation}. In
  \bibinfo{booktitle}{\emph{EMNLP}}.
\newblock


\bibitem[Xu et~al\mbox{.}(2020)]%
        {xu2020unsupervised}
\bibfield{author}{\bibinfo{person}{Shusheng Xu}, \bibinfo{person}{Xingxing
  Zhang}, \bibinfo{person}{Yi Wu}, \bibinfo{person}{Furu Wei}, {and}
  \bibinfo{person}{Ming Zhou}.} \bibinfo{year}{2020}\natexlab{}.
\newblock \showarticletitle{Unsupervised Extractive Summarization by
  Pre-training Hierarchical Transformers}. In
  \bibinfo{booktitle}{\emph{EMNLP}}.
\newblock


\bibitem[Yao et~al\mbox{.}(2021)]%
        {yao2021survey}
\bibfield{author}{\bibinfo{person}{Liuyi Yao}, \bibinfo{person}{Zhixuan Chu},
  \bibinfo{person}{Sheng Li}, \bibinfo{person}{Yaliang Li},
  \bibinfo{person}{Jing Gao}, {and} \bibinfo{person}{Aidong Zhang}.}
  \bibinfo{year}{2021}\natexlab{}.
\newblock \showarticletitle{A Survey on Causal Inference}.
\newblock \bibinfo{journal}{\emph{{ACM} Trans. Knowl. Discov. Data}}
  \bibinfo{volume}{15}, \bibinfo{number}{5} (\bibinfo{year}{2021}).
\newblock


\bibitem[Yeh et~al\mbox{.}(2005)]%
        {yeh2005text}
\bibfield{author}{\bibinfo{person}{Jen-Yuan Yeh}, \bibinfo{person}{Hao-Ren Ke},
  \bibinfo{person}{Wei-Pang Yang}, {and} \bibinfo{person}{I-Heng Meng}.}
  \bibinfo{year}{2005}\natexlab{}.
\newblock \showarticletitle{Text summarization using a trainable summarizer and
  latent semantic analysis}.
\newblock \bibinfo{journal}{\emph{Information processing \& management}}
  \bibinfo{volume}{41}, \bibinfo{number}{1} (\bibinfo{year}{2005}).
\newblock


\bibitem[Zajic et~al\mbox{.}(2004)]%
        {zajic2004bbn}
\bibfield{author}{\bibinfo{person}{David Zajic}, \bibinfo{person}{Bonnie Dorr},
  {and} \bibinfo{person}{Richard Schwartz}.} \bibinfo{year}{2004}\natexlab{}.
\newblock \showarticletitle{Bbn/umd at duc-2004: Topiary}. In
  \bibinfo{booktitle}{\emph{HLT-NAACL}}.
\newblock


\bibitem[Zeng et~al\mbox{.}(2019)]%
        {zeng2019you}
\bibfield{author}{\bibinfo{person}{Jichuan Zeng}, \bibinfo{person}{Jing Li},
  \bibinfo{person}{Yulan He}, \bibinfo{person}{Cuiyun Gao},
  \bibinfo{person}{Michael~R. Lyu}, {and} \bibinfo{person}{Irwin King}.}
  \bibinfo{year}{2019}\natexlab{}.
\newblock \showarticletitle{What You Say and How You Say it: Joint Modeling of
  Topics and Discourse in Microblog Conversations}.
\newblock \bibinfo{journal}{\emph{Transactions of the Association for
  Computational Linguistics}}  \bibinfo{volume}{7} (\bibinfo{year}{2019}).
\newblock


\bibitem[Zhang et~al\mbox{.}(2020)]%
        {zhang2020pegasus}
\bibfield{author}{\bibinfo{person}{Jingqing Zhang}, \bibinfo{person}{Yao Zhao},
  \bibinfo{person}{Mohammad Saleh}, {and} \bibinfo{person}{Peter~J. Liu}.}
  \bibinfo{year}{2020}\natexlab{}.
\newblock \showarticletitle{{PEGASUS:} Pre-training with Extracted
  Gap-sentences for Abstractive Summarization}. In
  \bibinfo{booktitle}{\emph{ICML}}, Vol.~\bibinfo{volume}{119}.
\newblock


\bibitem[Zhang et~al\mbox{.}(2021a)]%
        {zhang2021deep}
\bibfield{author}{\bibinfo{person}{Xingxuan Zhang}, \bibinfo{person}{Peng Cui},
  \bibinfo{person}{Renzhe Xu}, \bibinfo{person}{Linjun Zhou},
  \bibinfo{person}{Yue He}, {and} \bibinfo{person}{Zheyan Shen}.}
  \bibinfo{year}{2021}\natexlab{a}.
\newblock \showarticletitle{Deep Stable Learning for Out-of-Distribution
  Generalization}. In \bibinfo{booktitle}{\emph{CVPR}}.
\newblock


\bibitem[Zhang et~al\mbox{.}(2021b)]%
        {zhang2021causal}
\bibfield{author}{\bibinfo{person}{Yang Zhang}, \bibinfo{person}{Fuli Feng},
  \bibinfo{person}{Xiangnan He}, \bibinfo{person}{Tianxin Wei},
  \bibinfo{person}{Chonggang Song}, \bibinfo{person}{Guohui Ling}, {and}
  \bibinfo{person}{Yongdong Zhang}.} \bibinfo{year}{2021}\natexlab{b}.
\newblock \showarticletitle{Causal Intervention for Leveraging Popularity Bias
  in Recommendation}. In \bibinfo{booktitle}{\emph{SIGIR}}.
\newblock


\bibitem[Zheng et~al\mbox{.}(2021)]%
        {zheng2021disentangling}
\bibfield{author}{\bibinfo{person}{Yu Zheng}, \bibinfo{person}{Chen Gao},
  \bibinfo{person}{Xiang Li}, \bibinfo{person}{Xiangnan He},
  \bibinfo{person}{Yong Li}, {and} \bibinfo{person}{Depeng Jin}.}
  \bibinfo{year}{2021}\natexlab{}.
\newblock \showarticletitle{Disentangling User Interest and Conformity for
  Recommendation with Causal Embedding}. In \bibinfo{booktitle}{\emph{WWW}}.
\newblock


\bibitem[Zhong et~al\mbox{.}(2020)]%
        {zhong2020extractive}
\bibfield{author}{\bibinfo{person}{Ming Zhong}, \bibinfo{person}{Pengfei Liu},
  \bibinfo{person}{Yiran Chen}, \bibinfo{person}{Danqing Wang},
  \bibinfo{person}{Xipeng Qiu}, {and} \bibinfo{person}{Xuanjing Huang}.}
  \bibinfo{year}{2020}\natexlab{}.
\newblock \showarticletitle{Extractive Summarization as Text Matching}. In
  \bibinfo{booktitle}{\emph{ACL}}.
\newblock


\bibitem[Zhong et~al\mbox{.}(2019)]%
        {zhong2019searching}
\bibfield{author}{\bibinfo{person}{Ming Zhong}, \bibinfo{person}{Pengfei Liu},
  \bibinfo{person}{Danqing Wang}, \bibinfo{person}{Xipeng Qiu}, {and}
  \bibinfo{person}{Xuanjing Huang}.} \bibinfo{year}{2019}\natexlab{}.
\newblock \showarticletitle{Searching for Effective Neural Extractive
  Summarization: What Works and What{'}s Next}. In
  \bibinfo{booktitle}{\emph{ACL}}.
\newblock


\bibitem[Zhou et~al\mbox{.}(2018)]%
        {zhou2018neural}
\bibfield{author}{\bibinfo{person}{Qingyu Zhou}, \bibinfo{person}{Nan Yang},
  \bibinfo{person}{Furu Wei}, \bibinfo{person}{Shaohan Huang},
  \bibinfo{person}{Ming Zhou}, {and} \bibinfo{person}{Tiejun Zhao}.}
  \bibinfo{year}{2018}\natexlab{}.
\newblock \showarticletitle{Neural Document Summarization by Jointly Learning
  to Score and Select Sentences}. In \bibinfo{booktitle}{\emph{ACL}}.
\newblock


\bibitem[Zou et~al\mbox{.}(2022)]%
        {zou2022divide}
\bibfield{author}{\bibinfo{person}{Yicheng Zou}, \bibinfo{person}{Hongwei Liu},
  \bibinfo{person}{Tao Gui}, \bibinfo{person}{Junzhe Wang}, \bibinfo{person}{Qi
  Zhang}, \bibinfo{person}{Meng Tang}, \bibinfo{person}{Haixiang Li}, {and}
  \bibinfo{person}{Daniell Wang}.} \bibinfo{year}{2022}\natexlab{}.
\newblock \showarticletitle{Divide and Conquer: Text Semantic Matching with
  Disentangled Keywords and Intents}. In \bibinfo{booktitle}{\emph{ACL}}.
\newblock


\bibitem[Zou et~al\mbox{.}(2020)]%
        {zou2020pre}
\bibfield{author}{\bibinfo{person}{Yanyan Zou}, \bibinfo{person}{Xingxing
  Zhang}, \bibinfo{person}{Wei Lu}, \bibinfo{person}{Furu Wei}, {and}
  \bibinfo{person}{Ming Zhou}.} \bibinfo{year}{2020}\natexlab{}.
\newblock \showarticletitle{Pre-training for Abstractive Document Summarization
  by Reinstating Source Text}. In \bibinfo{booktitle}{\emph{EMNLP}}.
\newblock


\end{thebibliography}



\end{document}